\documentclass[]{fairmeta}

\def\benchmark{GenEval~2}
\def\gm{Soft-TIFA$_\text{GM}$}
\def\am{Soft-TIFA$_\text{AM}$}

\usepackage{bbm}
\usepackage{multirow}

\title{GenEval 2: Addressing Benchmark Drift in Text-to-Image Evaluation}

\author[1,2,3,*]{Amita Kamath}
\author[3]{Kai-Wei Chang}
\author[2,4]{Ranjay Krishna}
\author[1,2]{Luke Zettlemoyer}
\author[1,\dagger]{Yushi Hu}
\author[1,\dagger]{\\Marjan Ghazvininejad}

\affiliation[1]{FAIR at Meta}
\affiliation[2]{University of Washington}
\affiliation[3]{University of California, Los Angeles}
\affiliation[4]{Allen Institute for AI}

\contribution[*]{Work done at Meta}
\contribution[\dagger]{Joint last author}

\abstract{Automating Text-to-Image (T2I) model evaluation is challenging; a judge model must be used to score correctness, and test prompts must be selected to be challenging for current T2I models but not the judge.  
We argue that satisfying these constraints can lead to {\em benchmark drift} over time, where the static benchmark judges fail to keep up with newer model capabilities. 
We show that benchmark drift is a significant problem for GenEval, one of the most popular T2I benchmarks. 
Although GenEval was well-aligned with human judgment at the time of its release, it has drifted far from human judgment over time---resulting in an absolute error of as much as 17.7\% for current models. 
This level of drift strongly suggests that GenEval has been saturated for some time, as we verify via a large-scale human study. 
To help fill this benchmarking gap, we introduce a new benchmark, \benchmark, with improved coverage of primitive visual concepts and higher degrees of compositionality, which we show is more challenging for current models. 
We also introduce Soft-TIFA, an evaluation method for \benchmark~that combines judgments for visual primitives, which we show is more well-aligned with human judgment and argue is less likely to drift from human-alignment over time (as compared to more holistic judges such as VQAScore). 
Although we hope \benchmark~will provide a strong benchmark for many years, avoiding benchmark drift is far from guaranteed and our work, more generally, highlights the importance of continual audits and improvement for T2I and related automated model evaluation benchmarks. }

\date{\today}
\correspondence{Amita Kamath at \email{kamatha@cs.washington.edu}, Yushi Hu at \email{yushihu@meta.com}, \\Marjan Ghazvininejad at \email{ghazvini@meta.com}}

\metadata[Code]{\url{https://github.com/facebookresearch/GenEval2}}

\begin{document}

\maketitle

\section{Introduction}
\label{sec:introduction}

\begin{figure}[h!]
    \centering

    \begin{subfigure}{0.47\linewidth}
        \centering
        \includegraphics[width=0.75\linewidth]{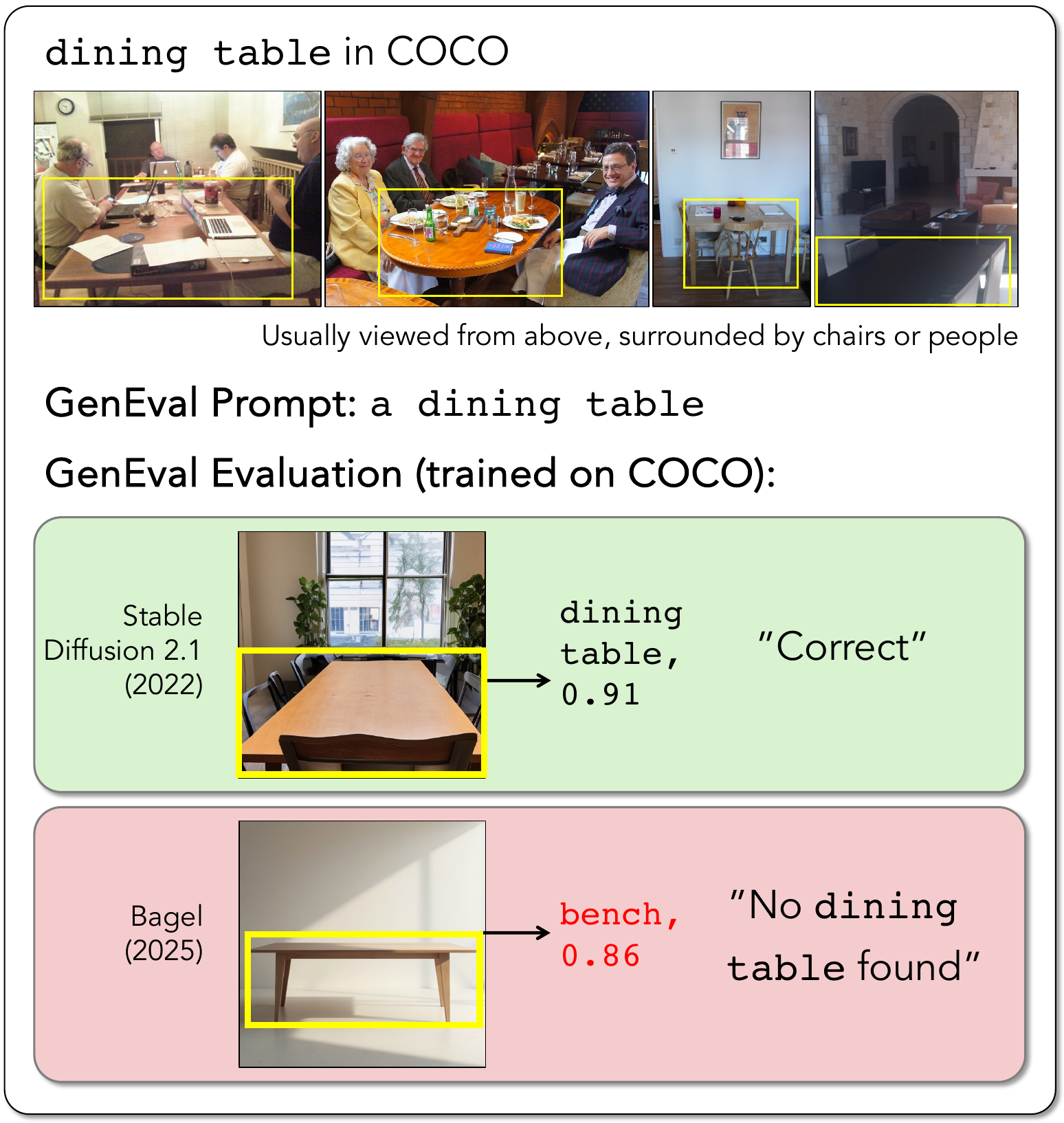}
        \caption{GenEval relies on CLIP~\citep{radford2021learning} and a detector trained on COCO~\citep{maskformer}, which are no longer reliable for evaluating recent T2I models. 
        }
        \label{fig:side_top}
    \end{subfigure}
    \hfill
    \begin{subfigure}{0.5\linewidth}
        \centering
        \includegraphics[width=\linewidth]{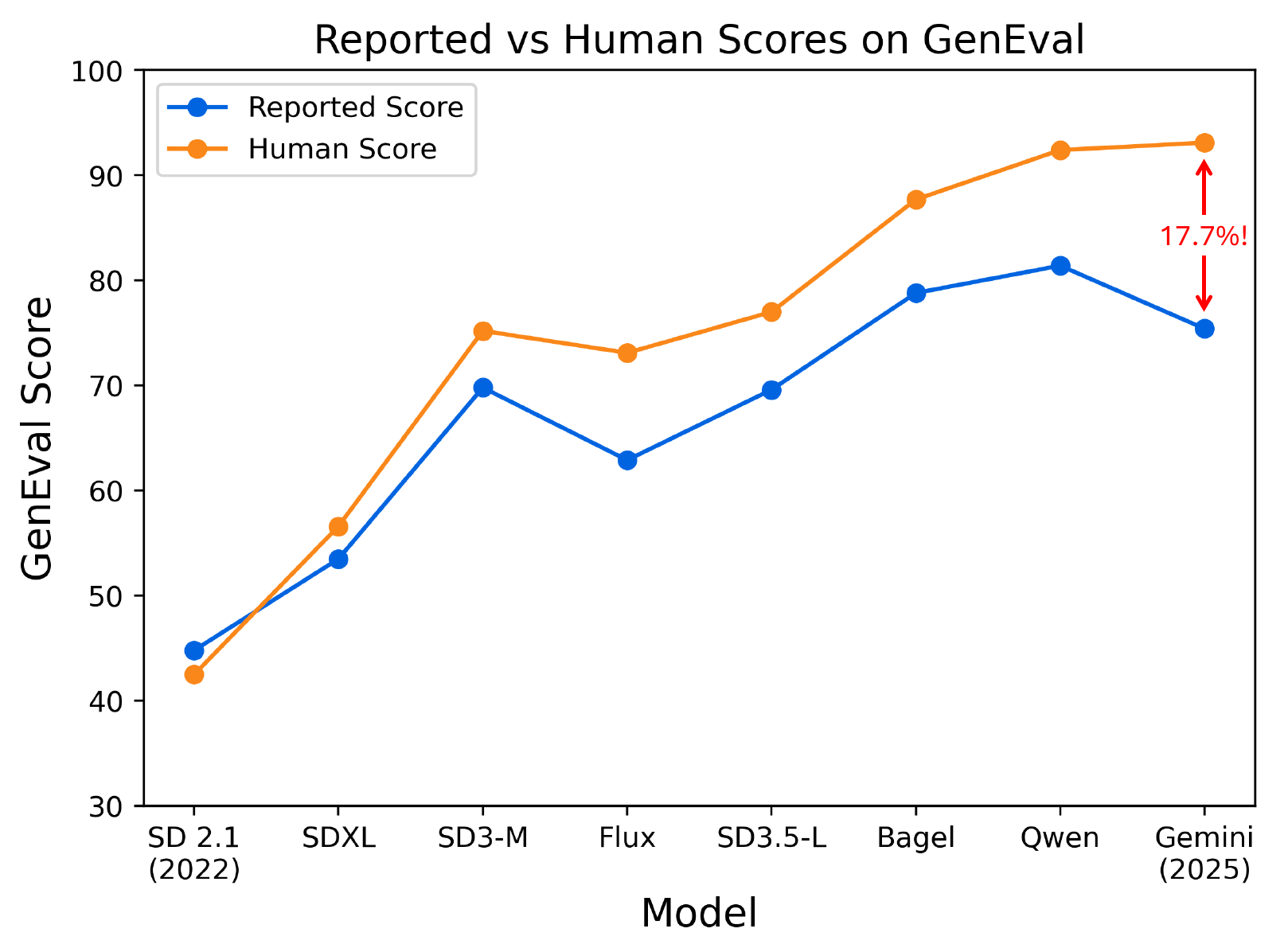}
        \caption{The gap between human and automatic evaluation scores on GenEval increases as T2I models become better, and eventually saturate the prompts.}
        \label{fig:side_bottom}
    \end{subfigure}

    \caption{With the distribution shift of Text-to-Image (T2I) models' outputs over time, we reveal that the model-based evaluation of GenEval decreases in human-alignment, masking the fact that the benchmark is now saturated. 
    We introduce \benchmark, a more robust benchmark that is challenging for state-of-the-art T2I models, alongside an evaluation method, Soft-TIFA, that is less likely to suffer benchmark drift. \vspace{-1.5em}
    }
    \label{fig:teaser}
\end{figure}

Text-to-Image (T2I) models are becoming increasingly capable \citep{deng2025emerging, wu2025qwen, flux2024, comanici2025gemini}, with models training on increasing amounts of natural and synthesized data. 
Their rapid progress has been both driven and measured by T2I benchmarks, for everything from basic capabilities like object colors and counts \citep{ghosh2023geneval, huang2023t2i, li2024genai} to advanced capabilities like knowledge and reasoning \citep{niu2025wise,chang2025oneig,sun2025t2i, chen2025r2i}. 
These benchmarks employ model-based evaluation: images generated by T2I models are evaluated using either a combination of specialized models such as object detectors and image-text matching models \citep{ghosh2023geneval, huang2023t2i}, or a single VQA model \citep{li2024genai,niu2025wise,chang2025oneig}. 

However, we raise a critical question: \textit{given how much T2I models' capabilities have changed, are the evaluations of longer-standing benchmarks still valid?} 
To investigate, we study one of the most prominent benchmarks to measure basic T2I capabilities: GenEval \citep{ghosh2023geneval}. This benchmark has been a primary evaluation in many popular T2I papers over the past three years including (but not limited to) 
Stable Diffusion 3 \citep{esser2024scaling},
Transfusion \citep{zhou2024transfusion},
Emu3 \citep{wang2024emu3},
Show-o \citep{xie2024show}, 
SEED-X \citep{ge2024seed}, 
MetaQueries \citep{pan2025transfer}, 
BAGEL \citep{deng2025emerging}, 
Janus \citep{wu2025janus}, 
OmniGen \citep{xiao2025omnigen}, 
BLIP3-o \citep{chen2025blip3}, and
Qwen-Image \citep{wu2025qwen}.

However, despite being so ubiquitously used in T2I research---usually with reports of gains of $\sim$2--3\% over previous state-of-the-art models---we find that GenEval results on recent models can diverge from human judgment by an absolute error of as much as 17.7\%! 
We conduct a large-scale human study, collecting over 23,000 annotations evaluating GenEval outputs of 8 popular T2I models that have released over the past three years. 
We show that although GenEval's evaluation aligned well with human judgment at the time of its release, with a distribution shift in images generated by T2I models over time, its evaluation has drifted farther from human judgment, resulting in large errors when evaluating recent models (c.f. Figure \ref{fig:teaser}). We refer to this phenomenon as \textit{benchmark drift}.
Further, according to human evaluation, GenEval is now saturated,
with Gemini 2.5 Flash Image~\citep{comanici2025gemini} achieving a score of 96.7\%. 

We introduce a new T2I benchmark, \benchmark. While maintaining GenEval's goal of evaluating T2I models' basic capabilities as well as its templated design, our benchmark has increased coverage of concepts, varying degrees of compositionality,
and enables capability-specific and compositionality-targeted analyses\footnote{The focus on compositionality in our benchmark 
goes hand-in-hand with
controllability: if a T2I model can understand the difference between and correctly generate both ``a brown dog and an black cat'' and ``a black dog and a brown cat'', it is inherently more user-controllable.
}.
We collect human annotations for the 8 T2I models on a subset 
of \benchmark, showing that models struggle with spatial relations, transitive verb relations, and counting. 
Further, model performance tends to drop as prompt compositionality increases. Overall, while models perform well on certain individual parts of the compositional prompt (referred to as ``atoms''), the top-performing model achieves only 35.8\% accuracy at the prompt-level, showing significant room for improvement.

We also introduce Soft-TIFA, an evaluation method tailored to \benchmark's templated and compositional structure. Soft-TIFA jointly estimates atom-level and prompt-level performance of T2I models by combining judgments for visual primitives, and on \benchmark~prompts,  it is more aligned with human judgment than popular automatic metrics such as VQAScore~\citep{lin2024evaluating} and TIFA~\citep{hu2023tifa}. Using Qwen3-VL-8B~\citep{yang2025qwen3} as the underlying VQA model, Soft-TIFA attains an AUROC of 94.5\%, compared to 92.4\% for VQAScore and 91.6\% for TIFA. Notably, it also outperforms GPT-4o-based VQAScore (91.2\% AUROC). Moreover, Soft-TIFA is inherently less susceptible to benchmark drift: whereas for a given VQA model, VQAScore-based estimates tend to diverge from human judgments over time after the model's release, Soft-TIFA remains consistently well aligned---potentially, breaking the prompt down into primitives causes distribution shift in the image to impact the underlying VQA model less. Soft-TIFA relies only on an open-source VQA model, rather than proprietary APIs.

In essence, we argue that model-based evaluations in the rapidly-changing research environment of T2I models need to be continually audited and updated to maintain their validity in light of benchmark drift---as we have done here for GenEval.
We release 
our new benchmark and evaluation method to support further research at \url{https://github.com/facebookresearch/GenEval2}.

\section{Related Work}
\label{sec:related_work}
\textbf{T2I Benchmarks.} Several T2I benchmarks evaluate basic capabilities such as object counts, colors, positions, and attribution, including GenEval \citep{ghosh2023geneval} and T2I-CompBench \citep{huang2023t2i}. Going beyond this to evaluate verbs and scene-level information are GenAI-Bench \citep{li2024genai}, TIFA-Bench \citep{hu2023tifa}, DSG-Bench \citep{cho2023davidsonian}, TIIF-Bench \citep{wei2025tiif}, Gecko \citep{wiles2025revisitingtexttoimageevaluationgecko}, and GenEval++ \citep{ye2025echo}. 
Other T2I benchmarks target advanced capabilities such as world knowledge and reasoning, including WISE \citep{niu2025wise}, R2I-Bench \citep{chen2025r2i},  T2I-ReasonBench \citep{sun2025t2i}, and OneIG-Bench \citep{chang2025oneig}. These benchmarks all utilize model-based judges, as discussed below.

\textbf{T2I Evaluation Methods.}
GenEval \citep{ghosh2023geneval} and T2I-CompBench \citep{huang2023t2i} use specialized judge models such as object detectors and image-text matching models to evaluate T2I models. 
However, these methods struggle with images that call for flexibility, such as spatial relations with unusual perspectives. 
Hence, the field has moved towards evaluation methods predicated on VQA models. In each of these methods, a question or set of questions is answered by a VQA model based on the generated image in order to evaluate it:
TIFA \citep{hu2023tifa} uses an LLM to break the prompt down into component questions;
DSG \citep{cho2023davidsonian} constructs a Davidsonian Scene Graph to obtain the questions;
VQAScore \citep{lin2024evaluating} uses a single question containing the prompt; and 
OneIG-Bench \citep{chang2025oneig} and R2I-Bench \citep{chen2025r2i} use an LLM to generate questions per prompt along with scoring rubrics.

We call for the research community to consider benchmark drift when designing T2I benchmarks and evaluation methods, and introduce a robust and challenging benchmark that is suitable for current frontier T2I models, alongside a new evaluation method less susceptible to benchmark drift.


\section{Auditing Benchmark Drift in GenEval}
We conduct a detailed study of GenEval to determine the risk in using long-standing model-based evaluations for T2I models, i.e., whether the evaluation method in this widely-used benchmark remains valid today.

\begin{table*}[t]
\centering

\begin{minipage}[]{0.58\textwidth}
\centering
\resizebox{\textwidth}{!}{%
\begin{tabular}{lcccccc}
\toprule
 & \multicolumn{3}{c}{\textbf{Without Rewriting}} & \multicolumn{3}{c}{\textbf{With Rewriting}} \\
 & Reported & Human & Net & Reported & Human & Net \\
 & Score & Score & Deviation & Score & Score & Deviation\\
\midrule
\multicolumn{2}{l}{\textit{Stable Diffusion Model Series}} & & & & & \\
SD 2.1          & 44.8 & 42.5 & -2.4 & -    & -    & -    \\
SD XL           & 53.5 & 56.6 & 3.1  & -    & -    & -    \\
SD 3-med*        & 69.8 & 75.2 & 5.4  & 79.2 & 85.9 & 6.7  \\
SD 3.5-large*    & 69.6 & 77.0 & 7.4  & 82.5 & 92.8 & 10.3 \\
\midrule
\multicolumn{2}{l}{\textit{State-of-the-Art T2I Models}} & & & & & \\
Flux.1-dev*      & 62.9 & 73.1 & 10.1 & 79.6 & 95.3 & 15.7 \\
Bagel+CoT       & 78.8 & 87.7 & 8.9  & 86.1 & 95.1 & 9.0  \\
Qwen-Image      & 81.4 & 92.4 & 11.0 & 87.2 & 94.8 & 7.6  \\
Gemini 2.5-F-I  & 75.4 & 93.1 & 17.7 & 82.1 & 96.7 & 14.6 \\
\bottomrule
\end{tabular}
}
\caption{ 
Reported Score, Human-Annotated Score, and Net Deviation of T2I models on GenEval. The benchmark is fast-approaching saturation, as shown by high Human Scores of T2I models With Rewriting. (* These models use CLIP, which truncates the prompt at 77 tokens. This truncates the rewritten prompts.) }
\label{tab:geneval1_results}
\end{minipage}%
\hfill%
\begin{minipage}[]{0.4\textwidth}
\centering
\includegraphics[width=\textwidth]{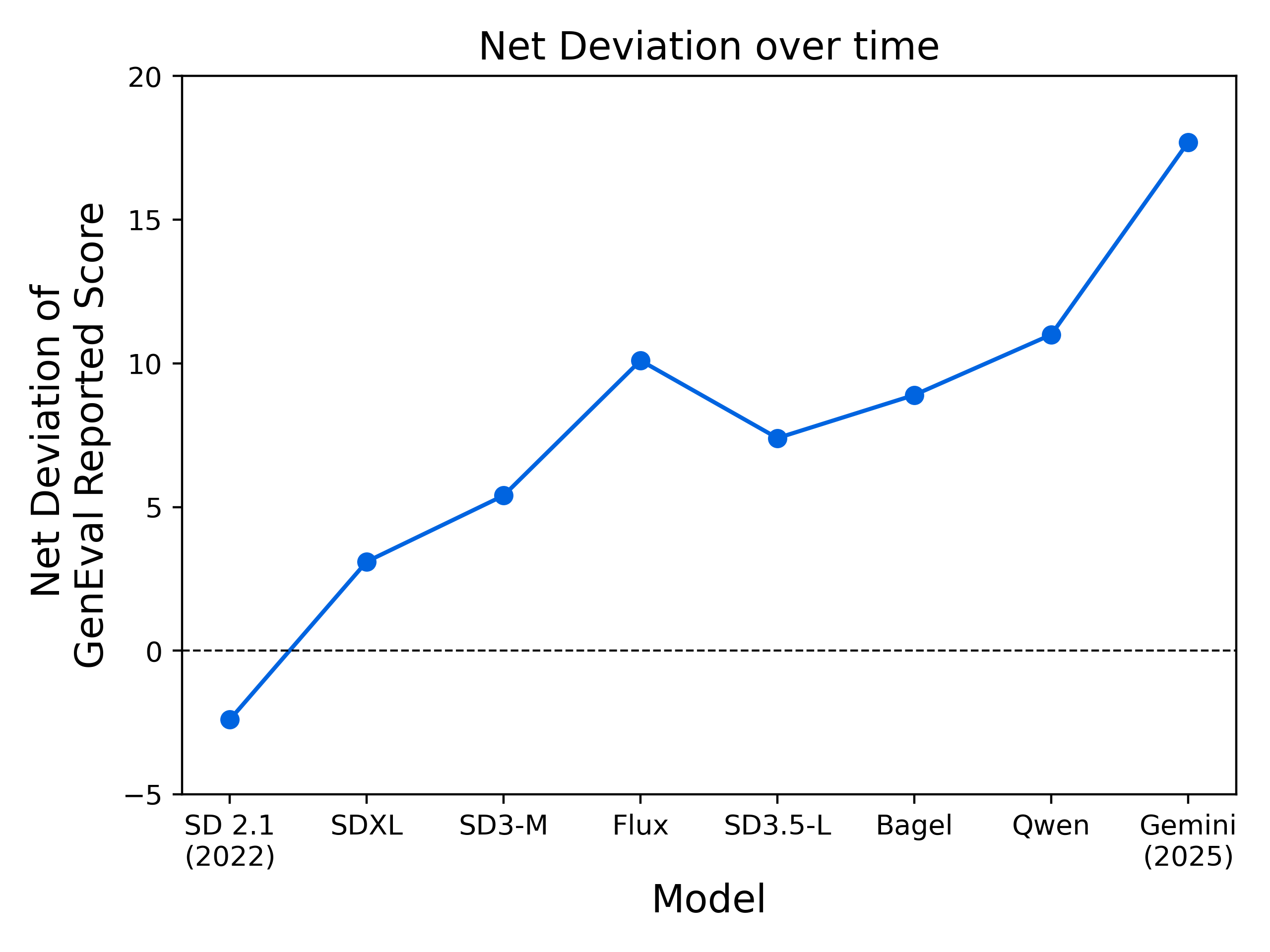}
\vspace{-2.3em}
\captionof{figure}{
Net deviation in reported score from human score on GenEval has increased significantly over time. The models on the X-axis are arranged by release date.
}
\label{fig:geneval1_deviation}
\end{minipage}

\end{table*}

\label{sec:geneval_1}
\subsection{Preliminary: GenEval Evaluation Method}
GenEval consists of 553 prompts, covering 6 categories evaluating T2I models' basic capabilities: One Object, Two Object, Color, Count, Position and Color Attribution. 
Objects are sourced from COCO categories \citep{lin2014microsoft}, colors are of 10 types, counts range from 2--4 and apply to a single object, the positions are in 2-dimensions (above, under, left of, and right of), and color attribution involves 2 objects.

The evaluation method varies across categories, but is model-based: a MaskFormer \citep{cheng2022masked} from MMDetection \citep{chen2019mmdetection}, trained on COCO \citep{lin2014microsoft} images, is used to identify objects in a generated image. 
Color is determined using CLIP \citep{radford2021learning} on the the object detector-identified bounding box after the background has been masked with the MaskFormer. Relative position between two objects is calculated mathematically based on their bounding box coordinates, after a buffer and maximum overlap is set between the two boxes.

This method has certain drawbacks, e.g.: it struggles to correctly mask the background of objects with holes, impacting color estimates; and relying on mathematical calculations for position suffers on examples such as ``suitcase under a table'', in which a correct image would likely be labeled as incorrect due to high bounding box overlap between the two objects.
However, 
it showed high human-alignment on the GenEval prompts at the time of its release, matching human judgment in 83\% of images generated by 
Stable Diffusion v2.1 \citep{rombach2022high}, IF-XL \citep{ifxl2023},
and LAION-5B with CLIP retrieval \citep{schuhmann2022laion}; 
increasing to 91\% on images with unanimous inter-annotator agreement.
As such, this benchmark was established as a standard evaluation, and is presented as a primary evaluation of T2I basic capabilities in major T2I papers over the past three years, including
Stable Diffusion 3 \citep{esser2024scaling},
Transfusion \citep{zhou2024transfusion},
Emu3 \citep{wang2024emu3},
Show-o \citep{xie2024show}, 
SEED-X \citep{ge2024seed}, 
MetaQueries \citep{pan2025transfer}, 
BAGEL \citep{deng2025emerging}, 
Janus \citep{wu2025janus}, 
OmniGen \citep{xiao2025omnigen}, 
BLIP3-o \citep{chen2025blip3}, and
Qwen-Image \citep{wu2025qwen}, among others.

\subsection{A New Human-Alignment Study of GenEval}
In order to determine whether GenEval has experienced benchmark drift, we conduct a large-scale human study, measuring the alignment of GenEval evaluations on 8 T2I models with human judgment. 
We collect human annotations for all 553 images in GenEval for each of these T2I models, in addition to images generated based on rewritten prompts for the 6 of these T2I models trained with rewriting techniques. Each image is annotated by 3 annotators, amounting to 23,226 human annotations.

\textbf{Models.} Our model selection covers two main criteria: we desire representation of the change in T2I models over time; in addition to state-of-the-art models capturing T2I capability today. For the former, we select the Stable Diffusion (SD) models: SD2.1 \citep{rombach2022high}, SDXL \citep{Podell2023SDXLIL}, SD3-medium \citep{esser2024scaling}, and SD3.5-large \citep{esser2024scaling}, released 2022-24. For the latter, we select Bagel (with thinking enabled) \citep{deng2025emerging}, Flux.1-dev \citep{flux2024}, Qwen-Image \citep{wu2025qwen} and Gemini 2.5 Flash Image \citep{comanici2025gemini}. 

\textbf{Rewriting techniques.} Recent T2I models employ rewriting techniques to improve performance \citep{BetkerImprovingIG, esser2024scaling, deng2025emerging}, first expanding the prompt to include more details, and then generating the image. As details behind rewriting methods have not been open-sourced, we use a rewritten version of GenEval prompts provided in the Bagel repository \citep{deng2025emerging}. We evaluate all T2I models with and without rewriting, except for SD2.1 and SDXL, which were not trained with rewriting. Note that SD3, SD3.5, and Flux.1-dev use CLIP \citep{radford2021learning}, which can only handle 77 text tokens for the prompt. This truncates the ending of many of the rewritten prompts, although the endings tend to discuss image-level stylistic details (e.g., ``...the composition is photographic in style...'') rather than information critical to prompt alignment, such as attributes or relations. To maintain consistency, we do not shorten the rewritten prompts, as they are in-line with those used by current state-of-the-art T2I models.

\textbf{Study details.} Annotators were provided with an image and two yes/no questions: (1) ``Does the image align with the prompt?''; and (2) ``Is the image of good quality? (i.e., are the objects well-formed?)''. While we only desire an answer to the first question for our analysis, we ask the annotators both questions to help them disentangle prompt alignment from image quality in their response. The instructions gave clear examples, along with discussions of edge cases, such as position-related prompts where the image captures an unusual perspective. Each data point was annotated by 3 annotators, and is labeled based on the majority vote. 85.1\% of the data points were labeled unanimously by the 3 annotators.
The instructions and user interface shown to the annotators are provided in Appendix \ref{sec:app_ge1_user_study}\footnote{The study was conducted via Invisible Tech \citep{invisible}.}.

\subsection{Findings from the Human Alignment Study of GenEval}
Having collected human labels for 8 T2I models' outputs on GenEval prompts, we now compare them to the benchmark's automated evaluation. Table \ref{tab:geneval1_results} contains the model-based score reported by GenEval, the human-judged score, and the net deviation between the two, for each T2I model. 

\textbf{GenEval is not well-aligned with human judgment on today's state-of-the-art models.}
GenEval scores are 11.0\% lower than human scores on average for today's state-of-the-art models, and as much as 17.7\% lower, in the case of Gemini 2.5 Flash Image. We note that this difference is very significant, given that many research papers report 
gains of 2-3\% over previous models. 

 \textbf{GenEval alignment with human judgment has decreased over time.} Figure \ref{fig:geneval1_deviation} shows the net deviation of GenEval reported scores from human scores across T2I models, arranged in chronological order of their release. While the deviation was within $\pm$3\% for SD2.1 and SDXL, which released before GenEval, it has steadily increased over the past 3 years, culminating in a deviation of 17.7\% for the latest T2I model, Gemini 2.5 Flash Image.

\textbf{Impact of rewriting techniques.} Across the 6 models trained with rewriting techniques, using rewritten prompts decreases human alignment for 4 models (SD3-med, SD3.5-large, Bagel and Flux.1-dev) by 2.5\% on average. However, it increases human alignment for Qwen-Image and Gemini 2.5 Flash Image by 3.3\% on average.

\textbf{GenEval is saturated.} Per the human score of various models on GenEval, depicted also in Figure \ref{fig:teaser}(b), all 4 state-of-the-art T2I models, released 2024-25, score 94.8\% and above with rewriting techniques, showing that GenEval is effectively saturated today, and has been for over a year\footnote{Flux.1-dev was released in August 2024.}.

\section{\benchmark: Addressing Benchmark Drift}
\begin{figure*}
    \centering
    \includegraphics[width=\linewidth]{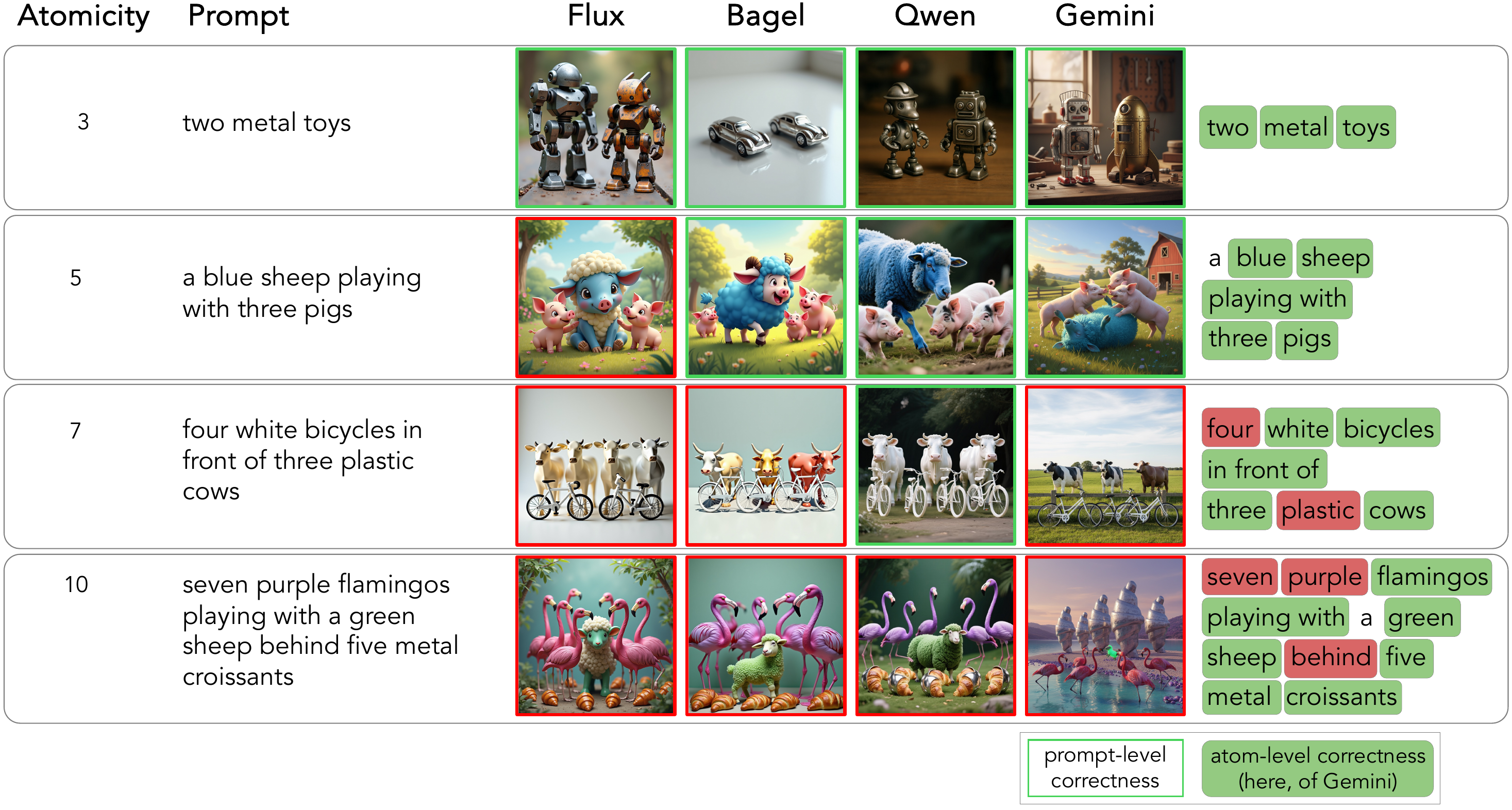}
    \caption{We present \benchmark, a T2I benchmark testing basic capabilities and increasing compositionality. Some samples are shown above. A prompt is considered correctly generated if \textit{all} component atoms are correctly generated.
    We show some samples of T2I model outputs on the benchmark, as well as prompt-level annotations for all models and atom-level annotations for Gemini 2.5 Flash Image.
    }
    \label{fig:ge2_placeholder}
\end{figure*}

Having highlighted the significant
benchmark drift in GenEval, i.e., 
that the distribution shift in T2I model outputs over time has resulted in a large decrease in human alignment and masked the benchmark's saturation, 
we present \benchmark, a novel, more robust, and challenging
benchmark to evaluate T2I models' basic capabilities. Our benchmark consists of a set of 800 prompts covering various skills, and at varying levels of compositionality. Figure \ref{fig:ge2_placeholder} depicts sample prompts and annotations from the benchmark. 

\label{sec:geneval_2}

\subsection{Benchmark Curation}
\label{sec:geneval_2_curation}

\textbf{Concepts and skills.} \benchmark~evaluates a similar-but-expanded set of skills compared to GenEval: it evaluates 40 unique objects (20 sourced from COCO, 20 otherwise), 18 attributes (colors, materials, and patterns), 9 relations (3D spatial prepositions and transitive verbs), and 6 counts (2--7). Of the 40 objects, 20 are animate (e.g., ``monkey'') and 20 are not (e.g., ``umbrella''). The 3D spatial prepositions include those in GenEval, as well as ``in front of'' and ``behind'', as in T2I-CompBench \citep{huang2023t2i}. We provide the full vocabulary of \benchmark~in Appendix \ref{sec:app_ge2_vocab}.

\textbf{Prompt structure.} Each prompt has 1--3 unique objects. We generate the prompts with templates, maintaining the process from GenEval, which allows for straightforward annotation, evaluation and analysis\footnote{This also prevents ambiguous prompts such as ``The rich, complex flavors of the aged wine tantalized the palate, a sensory feast of taste and smell.'' from T2I-CompBench \citep{huang2023t2i}, which was LLM-generated.}. 
The template follows the form 
\{count$_1$ or ``a''\}\{attribute$_1$\}\{object$_1$\}\{relation$_1$ or ``and''\}\{count$_2$ or ``a''\}\{attribute$_2$\}\{object$_2$\} for 2 objects, which can be extended to 3 objects. The counts, attributes, and relations are optional, resulting in varying degrees of compositionality, as discussed next.
The prompts combine the attributes, objects and relations at random, but for the transitive verb relation (e.g., ``chasing''), which is only applied between two animate objects; thus, our prompts contain unusual combinations, e.g., ``a green bagel to the left of a metal flamingo'', but not \textit{very} unusual, potentially anthropomorphizing ones, e.g. ``a green bagel chasing a metal flamingo'', maintaining focus on basic T2I capabilities. 

\textbf{Increasing compositionality.} We measure the compositionality of a prompt by measuring how many ``atoms'' it has, i.e., its atomicity, where an atom is an object, attribute, or relation: e.g., ``three pink pigs'' has 3 atoms, whereas ``three pink pigs jumping over a sheep'' has an atomicity of 5. We do not consider ``a'' and ``and'' to add to prompt compositionality. Our benchmark contains prompts with increasing compositionality ranging from 3 to 10 atoms, with 100 prompts of each atomicity, amounting to a total of 800 prompts. The atomicity of each prompt is noted in the benchmark, enabling compositionality-related analyses as shown in Section \ref{sec:ge2_analyses}. 

\textbf{Rewriting techniques.} We rewrite \benchmark~prompts following previous work, using GPT-4o \citep{hurst2024gpt}. 
While researchers are encouraged to use the rewriting technique of their choice, we provide 
the prompt given to GPT-4o to rewrite the benchmark in Appendix \ref{sec:app_ge2_rewriting}. 
As with GenEval, the rewritten prompts are longer than the text input window of CLIP, which is used in SD3, SD3.5-large and Flux.1-dev, resulting in truncation. However, here too, the information lost tends to be stylistic rather than prompt-critical.

\textbf{Evaluation.} We propose a new evaluation metric for \benchmark, Soft-TIFA, which is discussed in Section \ref{sec:evaluation_methods}.
For clarity, in this section, we discuss only results of the \textit{human annotations} of various T2I model outputs on \benchmark, i.e., human-judged scores, as collected in a human study and discussed in Section \ref{sec:ge2_human_annotation_results}. 

\textbf{Human study.} We collect human annotations for outputs from the 8 T2I models in Section \ref{sec:geneval_1} on a subset of \benchmark. We follow the same structure for the user study as with GenEval, but instead of requesting a yes/no answer for the question ``Does the image align with the prompt?'', we provide a checklist for each atom in the prompt for the user to select based on alignment of the image. Thus, we obtain atom-level annotations of alignment for each generated image, as shown in Figure \ref{fig:ge2_placeholder} for Gemini 2.5 Flash Image.
The instructions and user interface are provided in Appendix \ref{sec:app_ge2_user_study}.
Each image is annotated by 1 annotator, as we found high inter-annotator agreement in our pilot study of 200 data points (91\% unanimous scoring across 3 annotators).
We label 50\% of the benchmark in this manner, evenly split across models and rewriting methods\footnote{This study was also conducted by Invisible Tech \citep{invisible}.}.

\subsection{Human-Judged Scores on \benchmark}
\label{sec:ge2_human_annotation_results}
Having collected human-judged correctness across the 8 T2I models from Section \ref{sec:geneval_1} on \benchmark, 
we report 
atom-level and prompt-level results in Table \ref{tab:geneval2_human_results}, with and without rewriting. At the prompt-level, we present a binary measure of correctness, as in GenEval, where a prompt is considered correct only if each atom in the prompt has been labeled as correct. 

At the atom-level, the latest models perform quite well, with the highest-performing models achieving a human-judged score of 85.3\%. However, at the prompt-level, the highest-performing model achieves a score of only 35.8\%. Keeping in mind that \benchmark~evaluates only basic T2I capabilities, these results show that there remains significant room for improvement on compositional prompts.

\textbf{Rewriting techniques improve performance.} As shown in Table \ref{tab:geneval2_human_results}, rewriting the prompt aids T2I performance at both the atom-level and prompt-level across all models (in the second decimal for Gemini 2.5 Flash Image). 

\begin{table}[t]
\centering
\resizebox{0.65\linewidth}{!}{%
\begin{tabular}{lcccc}
\toprule
 & \multicolumn{2}{c}{\textbf{Without Rewriting}} & \multicolumn{2}{c}{\textbf{With Rewriting}} \\
 & Atom- & Prompt- & Atom- & Prompt-\\
 & level & level & level & level \\
 \midrule
 \multicolumn{1}{l}{\textit{Stable Diffusion Model Series}} & & \\
 SD 2.1 & 33.7 & 2.5 & - & -\\
 SD XL	& 45.3& 4.3 & - & -\\
 SD 3-med*	& 64.5  & 14.0 & 69.3 & 15.8\\
 SD 3.5-large* & 67.8 & 14.3 & 69.7 & 19.0\\
 \midrule
 \multicolumn{1}{l}{\textit{State-of-the-Art T2I Models}} & & \\
 Flux.1-dev*  &	67.0 & 15.3 & 80.4 & 29.5\\
 Bagel+CoT  & 71.7 & 16.0 & 76.8 & 22.3\\
 Qwen-Image  &	82.0 & 26.8& 85.3 & 35.8\\
 Gemini 2.5 Flash Image &84.4 & 31.0& 84.4 & 32.3\\
\bottomrule
\end{tabular}
}
\caption{
Human evaluation of \benchmark. While models show good performance at the atom-level, there is significant room for improvement at the prompt-level, even with rewriting techniques, highlighting the need for compositionality. (* These models use CLIP, which truncates the prompt at 77 tokens. This truncates the rewritten prompts.)
}
\label{tab:geneval2_human_results}
\end{table}

\subsection{Analyses of Capability and Compositionality}
\label{sec:ge2_analyses}

\begin{figure*}[t]
    \centering
    \begin{subfigure}[t]{0.48\linewidth}
        \centering
        \includegraphics[width=0.9\linewidth]{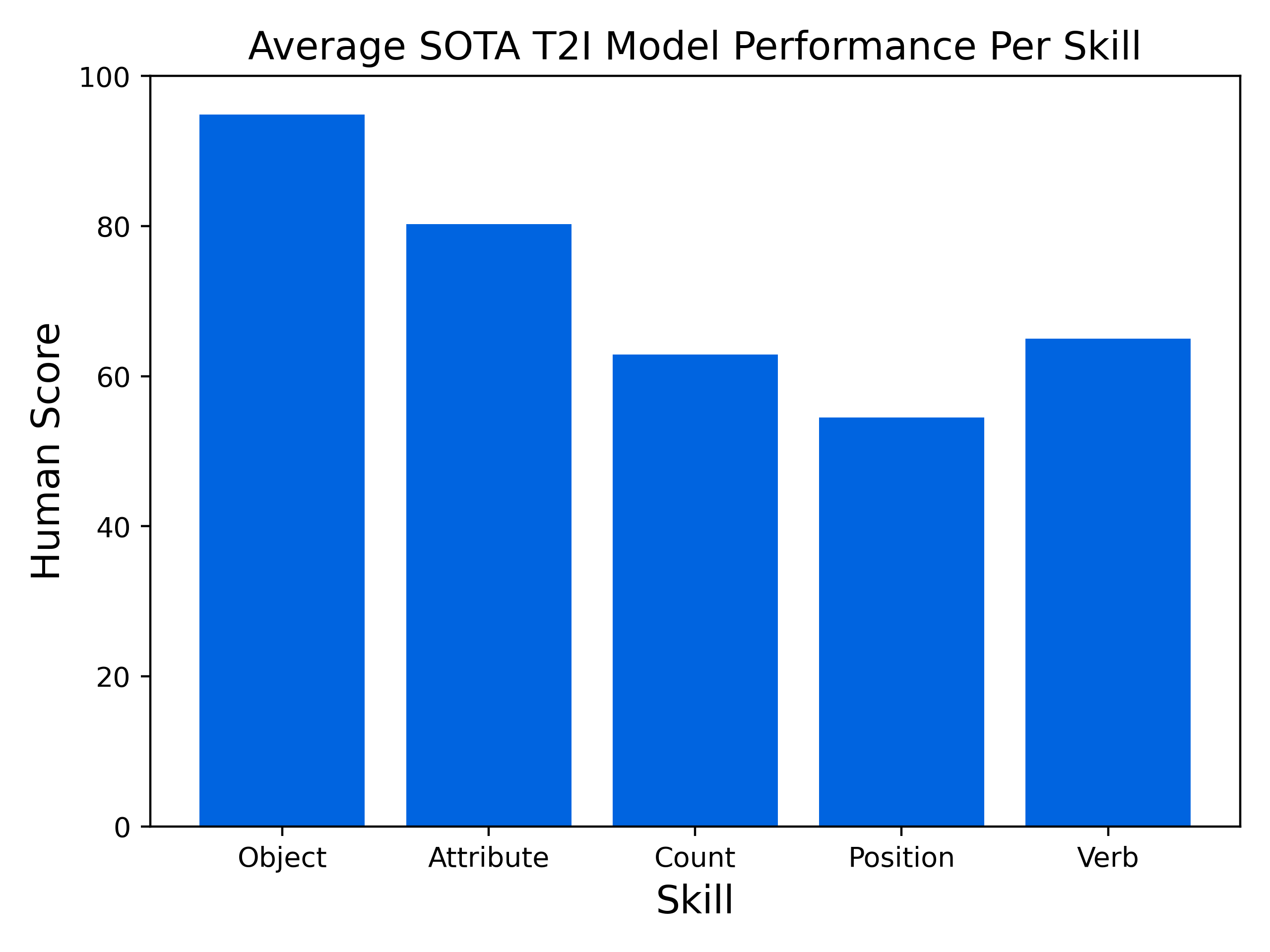}
        \caption{Average Human Score of SOTA T2I models per skill in \benchmark.}
        \label{fig:geneval2_skill}
    \end{subfigure}
    \hfill
    \begin{subfigure}[t]{0.48\linewidth}
        \centering
        \includegraphics[width=0.9\linewidth]{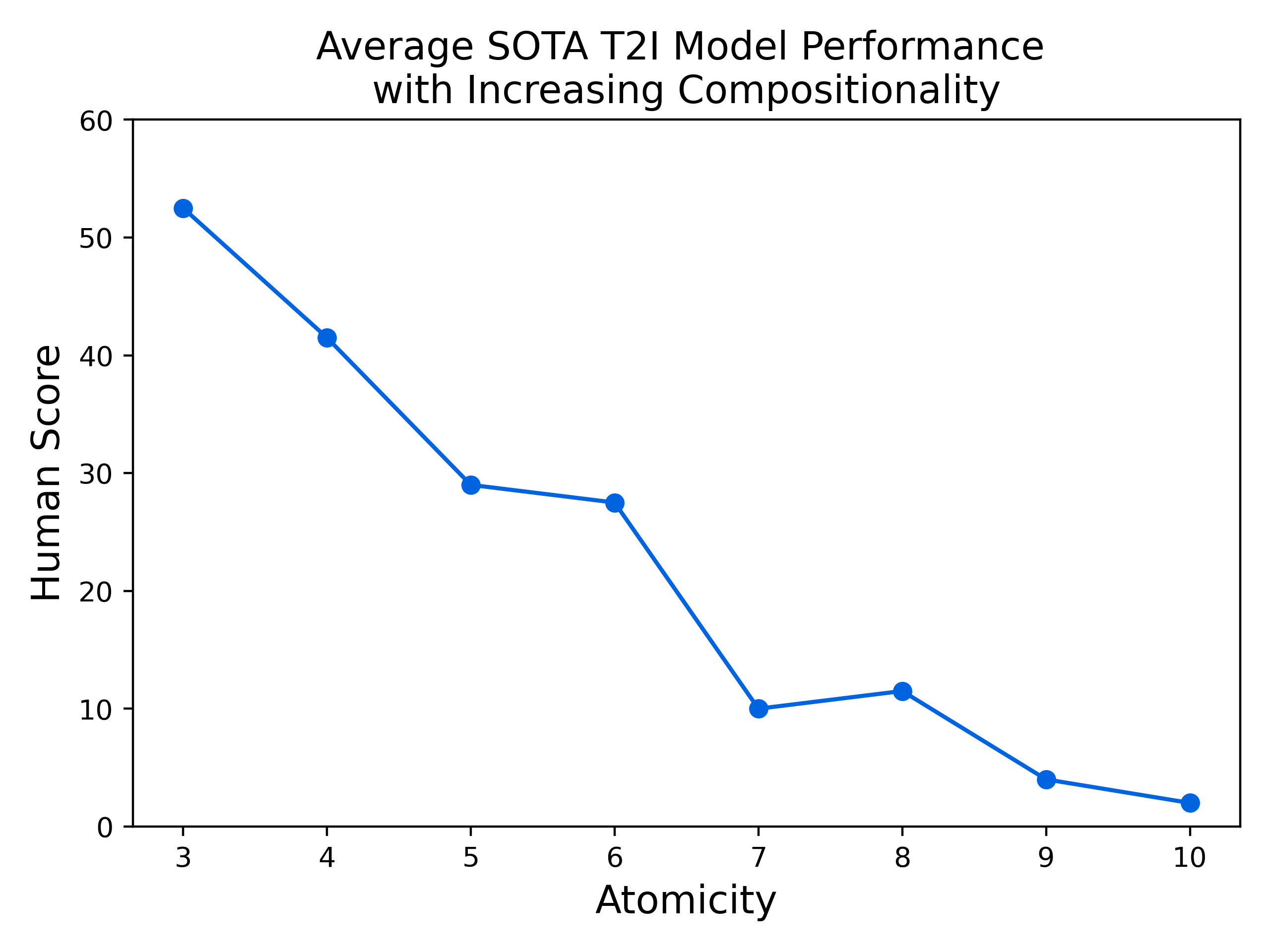}
        \caption{Average Human Score of SOTA T2I models for each level of compositionality (``atomicity'') in \benchmark.}
        \label{fig:geneval2_compositionality}
    \end{subfigure}
    \caption{\benchmark~enables various analyses of T2I models: (a) while state-of-the-art (SOTA) T2I models perform well at generating objects, and quite well at assigning them attributes, they struggle with counting, spatial relations, and transitive verb relations; (b) SOTA T2I model performance drops sharply as prompts become more complex. Per-model analyses are provided in Appendix \ref{sec:app_per_model}.}
    \label{fig:two_side_by_side}
\end{figure*}

\benchmark~was designed to encourage various types of analyses. Each prompt is labeled with its atomicity, as well as which skills\footnote{We use ``skill'' interchangeably with ``capability'' in this paper.} it requires (e.g., counting, spatial relations), enabling composition-related and skill-related analyses. In this section, we conduct these analyses on the 4 state-of-the-art T2I models discussed above (Flux.1-dev, Bagel, Qwen-Image and Gemini 2.5 Flash Image) to highlight patterns in performant models. An analysis of each of the 8 models is presented in Appendix \ref{sec:app_per_model}.

\textbf{Models particularly struggle with spatial relations, transitive verb relations, and counting.} Models' average per-skill performance on \benchmark~is shown in Figure \ref{fig:geneval2_skill}. Clearly, while they can generate various objects and attributes well, counting, spatial relations and transitive verbs remain challenging. This is consistent with findings from earlier work \citep{hu2023tifa, li2024genai}. Further, many of the attribute errors are caused by \textit{attribution} errors, when multiple attributes applied to different objects are present in the prompt. 

\textbf{Model performance drops as prompt compositionality increases.} On average, model performance decreases with increasing prompt compositionality, as shown in Figure \ref{fig:geneval2_compositionality}---at the highest levels of complexity, T2I model performance is close to 0. Clearly, state-of-the-art T2I models continue to struggle with more compositional prompts, which are central to many practical applications \citep{li2024genai}. 

Our benchmark design provides a detailed diagnosis for each model, highlighting its strengths and weaknesses along multiple axes: skills, compositionality, and rewriting (which goes hand-in-hand with prompt length), enabling targeted research in improving specific model capabilities: e.g., Qwen-Image achieves 99\% accuracy at generating specific objects, but only 60\% accuracy at relative positions of the objects (c.f. Appendix \ref{sec:app_per_model}).

\section{A New Evaluation Method for \benchmark: Soft-TIFA}
\label{sec:evaluation_methods}
Alongside our new benchmark \benchmark, we propose Soft-TIFA, a new evaluation method uniquely suited to the template-based and compositional nature of \benchmark. 

\subsection{Preliminaries: VQAScore and TIFA}
Soft-TIFA draws inspiration from VQAScore \citep{lin2024evaluating} and TIFA \citep{hu2023tifa}, two popular T2I evaluation methods predicated on VQA models. 

\textbf{VQAScore.} This method assigns a soft (continuous) score to each generated image,
that has been shown to align well with human judgment on GenAI-Bench \citep{li2024genai}. The per-image score is calculated under a VQA model $M$ as:
\begin{equation}
\begin{aligned}
\text{score} = P_{M}(&\text{``Yes'' } \mid \text{ image}, \\
&\text{``Does this image show \{prompt\}?}\\
&\text{Answer in one word, Yes or No.''})
\end{aligned}
\end{equation}

\textbf{TIFA.} This method generates question-answer pairs based on each prompt using an LLM. 
For an image with $N$ generated question-answer pairs where answers are denoted as $a_i$ and answers predicted by a VQA model based on the image are denoted as $\hat{a}_i$, the per-image score is:
\begin{equation}
\begin{aligned}
\text{score} = \frac{1}{N} \sum_{i=1}^{N}\mathbbm{1}[\hat{a}_i==a_i]
\end{aligned}
\end{equation}

\textbf{Note about rewriting techniques.} We note that rewriting techniques are only used for image generation, \textit{not} in evaluation: i.e., a T2I model sees the rewritten prompt to generate an image, but the evaluation compares that image against the original prompt, for all evaluation methods.

\subsection{Soft-TIFA}
\benchmark~is highly compositional: each prompt contains up to 10 atoms (objects, attributes or relations). We believe a TIFA-like approach would work well to evaluate the correctness of the generated image on each atom in the prompt. Further, \benchmark~is generated with templates: hence, the TIFA-like questions could also be generated with templates, ensuring full coverage over the prompt---a problem for TIFA itself with more compositional prompts, as it uses an LLM to generate the questions.
However, we also desire the soft score aspect of VQAScore, which allows the metric to capture VQA model uncertainty for each question. 

Hence, we present Soft-TIFA, which generates one question per atom of each prompt using the same templates used to generate \benchmark, and assigns \textit{soft scores} to each answer as predicted by the VQA model when given the generated image: i.e., for an image with $N$ generated question-answer pairs (where prompt atomicity is $N$), where answers are denoted $a_i$, the per-image score under a VQA model $M$ is:
\begin{equation}
\begin{aligned}
\text{score} = \text{mean}\{P_M(a_i)\}_{i=1}^N
\end{aligned}
\end{equation}

When the mean function across predictions for an image is the arithmetic mean, this method is referred to as \textbf{\am}, and captures \textit{atom-level} correctness of the T2I model. 
When the mean function is the geometric mean, this method is referred to as \textbf{\gm}, and captures \textit{prompt-level} correctness of the T2I model, as it penalizes low atom-level scores at the prompt-level (similar to our human prompt-level score calculation, where a prompt is incorrect if any component atom is incorrect). 
Both \am~and \gm~are averaged arithmetically across the benchmark to obtain a final score. 

That Soft-TIFA captures both atom-level and prompt-level correctness is
an important consideration when estimating T2I model performance on the compositional prompts in \benchmark. 
We use Qwen3-VL \citep{yang2025qwen3} as our VQA model, as it is both performant and open-source. 

\subsection{Human Alignment of Soft-TIFA on \benchmark}
We next determine whether Soft-TIFA is well-aligned with human judgment on \benchmark. We compare \am~and \gm~to existing evaluation methods, using AUROC to measure human alignment, as in VQAScore \citep{lin2024evaluating}. The human scores compared against are prompt-level scores, collected as discussed in Section \ref{sec:geneval_2_curation}.

As shown in Table \ref{tab:geneval2_auroc}, \am~matches or outperforms existing methods on rewritten and combined sets of prompts, and \gm~outperforms existing methods on the original, rewritten, and combined sets of \benchmark~prompts.
We attribute Soft-TIFA's high human-alignment to its structure, which is inherently designed to capture the varying levels of compositionality of T2I input prompts present in \benchmark. 

\begin{figure}[t] 
    \centering

    \begin{subfigure}[b]{0.49\columnwidth}
        \centering
        \includegraphics[width=\linewidth]{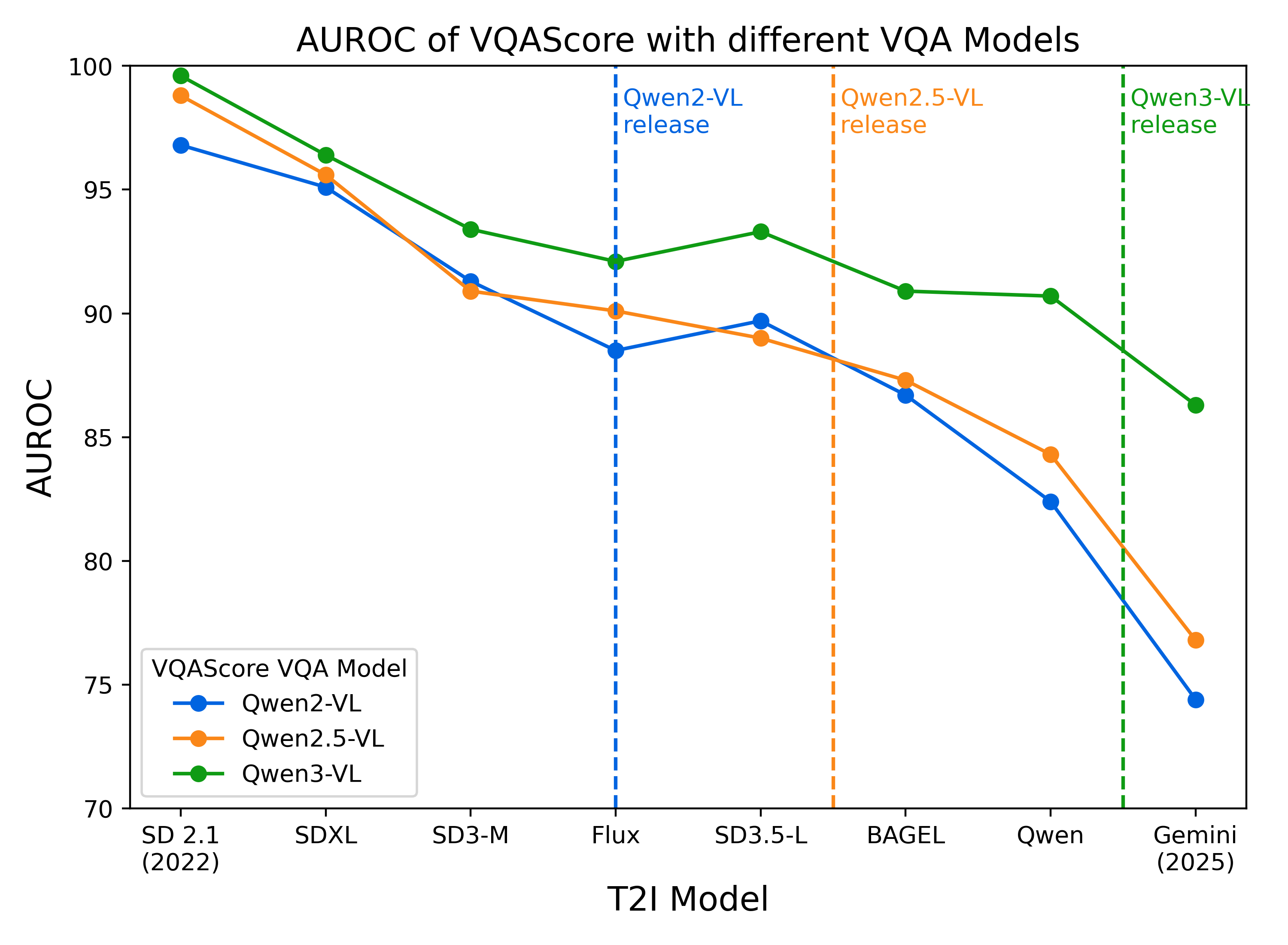}
        \caption{Human alignment of VQAScore on \benchmark~under different VQA models.}
        \label{fig:vqascore_alignment}
    \end{subfigure}
    \hfill
    \begin{subfigure}[b]{0.49\columnwidth}
        \centering
        \includegraphics[width=\linewidth]{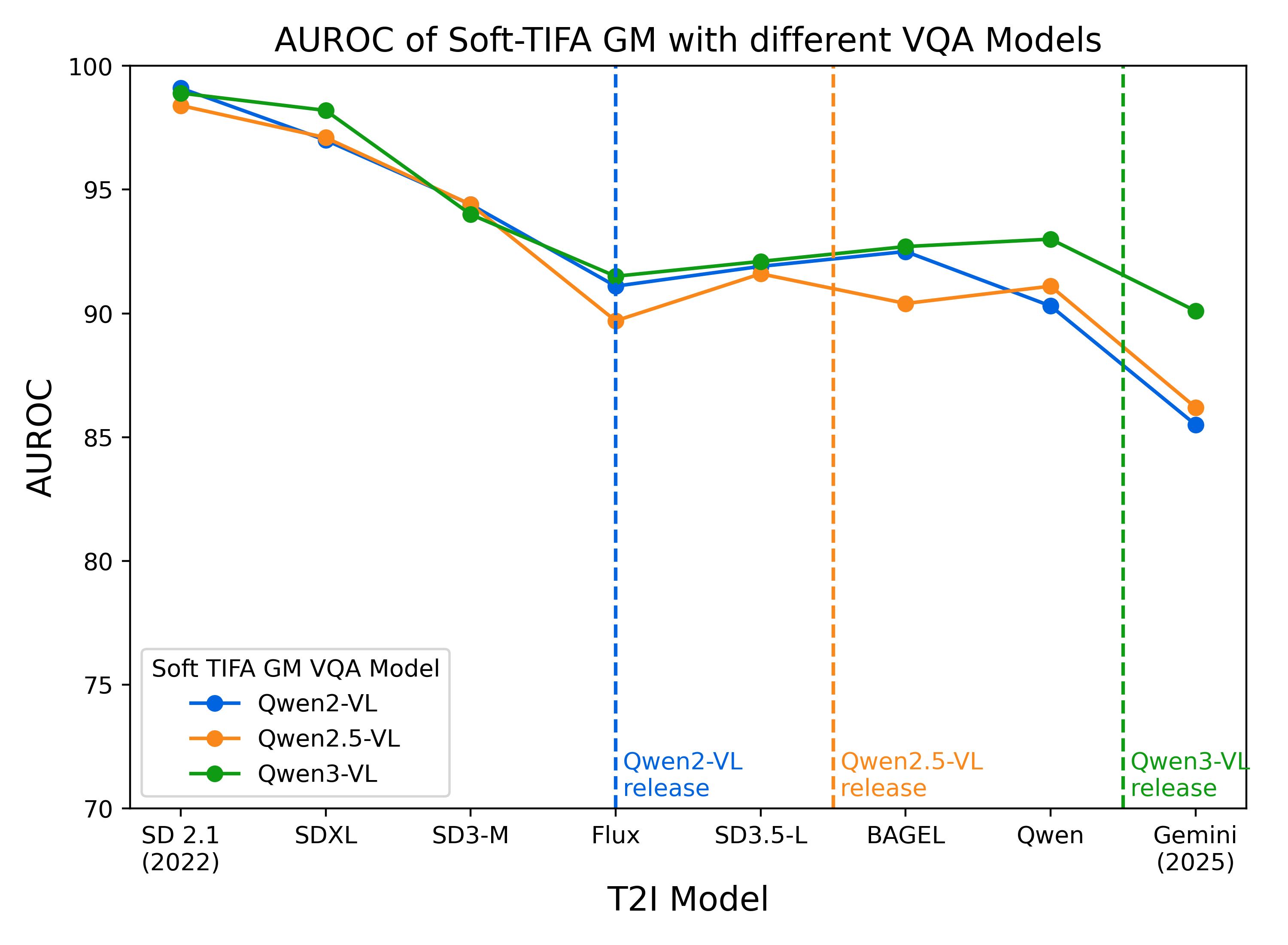}
        \caption{Human alignment of \gm~on \benchmark~under different VQA models.
        }
        \label{fig:softtifagm_alignment}
    \end{subfigure}

    \caption{Under all VQA models, \gm~achieves higher human alignment on \benchmark~across T2I models than VQAScore. Further, it is more robust to T2I output distribution shift over time, potentially because breaking the prompt into per-atom questions renders the VQA model more robust to image distribution shift in T2I outputs.
    }
    \vspace{-1em}
    \label{fig:metric_alignment}
\end{figure}

We also run this study with other strong, open-source VQA models underlying the evaluation methods. As presented in Figure \ref{fig:metric_alignment} and Appendix \ref{sec:app_different_vqa}, we find that our results from Table \ref{tab:geneval2_auroc} hold across Qwen2-VL \citep{wang2024qwen2} and Qwen2.5-VL \citep{bai2025qwen2}, and further outperform VQAScore based on GPT-4o. Hence, our method is not oversensitive to the underlying VQA model.



\begin{table}[t]
\centering
\resizebox{0.5\linewidth}{!}{%
\renewcommand{\arraystretch}{1.1}
\begin{tabular}{lccc}
\toprule
\textbf{Method} & \textbf{Original} & \textbf{Rewritten} & \textbf{Both} \\
\midrule
TIFA & 91.2 & 91.2 & 91.6 \\
VQAScore & \underline{93.5} & 90.5 & 92.4 \\
\am & 92.8 & \underline{92.9} & \underline{93.0}\\
\gm & \textbf{94.2} & \textbf{94.4} & \textbf{94.5} \\
\bottomrule
\end{tabular}
}
\caption{
AUROC of Qwen3-VL-based evaluation methods on \benchmark~: \gm~shows highest human-alignment across all sets of prompts. (\textbf{bold}=highest, \underline{underlined}=second-highest) 
\vspace{-1em}
}
\label{tab:geneval2_auroc}
\end{table}

\textbf{Per-skill and per-compositionality.} We calculate Soft-TIFA scores per-skill and per-compositionality to determine whether they reflect the human scores presented in Section \ref{sec:ge2_analyses}. As skill is an atom-level concept and compositionality is prompt-level, we use in our analyses \am~and \gm~respectively. We find that the trends are largely mirrored for both skills and levels of compositionality, with T2I models performing well at objects and attributes but not counting, verbs or positions, and T2I model performance dropping steeply with increase in prompt compositionality. The lone exception is the ``verb'' skill, in which VQA models are stricter than human judgment---showing room for improvement in VQA model judgment of more nuanced relations. Detailed results are shown in Appendix \ref{sec:app_soft_tifa_per}.

\subsection{Comparison of T2I Models}
\label{sec:ge2_model_comparison_tifa}
Table \ref{tab:geneval2_soft_tifa_results} shows the Soft-TIFA scores of our 8 T2I models on \benchmark, alongside the human-judged scores from Section \ref{sec:ge2_human_annotation_results}. While the four state-of-the-art T2I models perform well at the atom-level (as shown by both the human atom-level and \am~scores), they have significant scope for improvement at the prompt-level (as shown by the human prompt-level and \gm~scores). Soft-TIFA largely retains the ranking of human-scored T2I models.

\begin{table}[t]
\centering
\resizebox{0.78\linewidth}{!}{%
\begin{tabular}{lcccc}
\toprule
\multirow{2}{*}{\textbf{Model}} & \textbf{Human Score} & \textbf{Soft}\hspace{15px} & \textbf{Human Score} & \textbf{Soft} \hspace{15px} \\
 & \textbf{(Atom-level)} & \textbf{TIFA$_\text{AM}$} & \textbf{(Prompt-level)} & \textbf{TIFA$_\text{GM}$} \\
\midrule
 \multicolumn{1}{l}{\textit{Stable Diffusion Model Series}} & & \\
SD 2.1        & 33.7 & 40.8 & 2.5  & 5.2 \\
SDXL          & 45.3 & 50.1 & 4.3  & 9.1 \\
SD3           & 64.5 & 66.3 & 14.0 & 21.3 \\
SD3.5-large   & 67.8 & 69.7 & 14.3 & 22.8 \\
\midrule
 \multicolumn{1}{l}{\textit{State-of-the-Art T2I Models}} & & \\
Flux          & 67.0 & 67.1 & 15.3 & 21.1 \\
Bagel+CoT     & 71.7 & 70.9 & 16.0 & 23.1 \\
Qwen          & 82.0 & 80.8 & 26.8 & 33.8 \\
Gemini        & 84.4 & 82.8 & 31.0 & 44.6 \\
\bottomrule
\end{tabular}
}
\caption{Soft-TIFA evaluation of T2I models on \benchmark, compared against human-judged scores from Table \ref{tab:geneval2_human_results}. \am~captures atom-level correctness of the T2I model, and \gm~captures prompt-level correctness. 
}
\label{tab:geneval2_soft_tifa_results}
\end{table}

\subsection{Evaluation Methods vs. Benchmark Drift}
\label{sec:which_model}
We select Qwen3-VL as the VQA model underlying Soft-TIFA. However, we highlight the need to audit and update evaluation components as time passes due to benchmark drift with shifts in T2I output image distribution, as studied in this paper.
To underscore our point, we experiment with three VQA models released at different points of time: Qwen2-VL (released in August 2024), Qwen2.5-VL (released in January 2025) and Qwen3-VL (released in September 2025). We calculate the AUROC of VQAScore and \gm~when based on each of these VQA models for each of the 8 T2I models, to estimate human-alignment on \benchmark~based on our human scores from Section \ref{sec:ge2_human_annotation_results}.

As shown in Figure \ref{fig:metric_alignment} (a), with the VQAScore metric,
each of the 3 VQA models achieves a high AUROC for the T2I models that came out before their releases, \textit{then achieve increasingly lower AUROC for the T2I models that came out afterward.}
Clearly, the distribution shift in T2I model output after each VQA model's release could not be accounted for, and thus caused a decrease in human-alignment of the VQAScore evaluation.

With the \gm~metric in Figure \ref{fig:metric_alignment} (b), this phenomenon appears to be mitigated, as shown by less of a drop in human-alignment AUROC after each VQA model's release---potentially, breaking the prompt down into per-atom questions causes distribution shift in the image to impact the VQA model less. However, it is entirely possible that greater distribution shift over a longer period of time may cause \gm~to show the same drift as VQAScore.

Our results show that merely shifting away from specialized models such as object detectors for T2I evaluation (as in GenEval) towards VQA-based methods (as in GenAI-Bench) is insufficient to address benchmark drift.
T2I models' output image distribution will keep changing, and evaluation methods need to change along with them to keep up. 

\textbf{A note about GPT models for evaluation.}
Many recent T2I benchmarks utilize GPT-4o for T2I evaluation, including WISE \citep{niu2025wise}, R2I-Bench \citep{chen2025r2i}, TIIF-Bench \citep{wei2025tiif}, and
DreamBench++ \citep{peng2024dreambench++}, among others. 
However, GPT-4o is closed-source and constantly changing, rendering results difficult to reproduce---not to mention that the model could be deprecated\footnote{As it has been in the past: Sam Altman, X (formerly Twitter), Aug. 8 2025, \url{https://x.com/sama/status/1953506023391592494}}.
To account for this in addition to benchmark drift, we encourage the research community to utilize strong open-source VLMs instead of closed-source models.







\section{Conclusion}
\label{sec:conclusion}
In this work, we study benchmark drift of T2I evaluations. 
We find that the output distribution of T2I models has shifted so greatly over the past 3 years that GenEval, one of the most popular T2I benchmarks to evaluate basic capabilities, is no longer well-human-aligned. 
We quantify the benchmark drift of GenEval with a large-scale human study across 8 T2I models, finding that results on recent models can diverge from human judgment by an absolute error of as much as 17.7\%. 
Further, according to the human evaluation, GenEval is now saturated. 
To help bridge this gap, we introduce \benchmark, a challenging new T2I benchmark inspired by GenEval that evaluates basic capabilities on increasingly compositional prompts, and 
allows for targeted analyses. Human annotations of T2I models on \benchmark~show that state-of-the-art T2I models still have significant room for improvement.
We also introduce Soft-TIFA, an evaluation method designed for \benchmark~that provides estimates of atom-level as well as prompt-level performance of T2I models. Soft-TIFA has higher human alignment than VQAScore and TIFA, and is more robust against benchmark drift.
Our work serves as a call for caution in the light of benchmark drift, and underscores the need to continually audit and update T2I evaluation benchmarks.

\section*{Acknowledgments}
We thank Jiahui Chen and Melissa Hall for interesting discussions and helpful feedback on our work, as well as Jonea Gordon and Vanessa Stark. We also thank Cynthia Gao and Justin Hovey, and the team at Invisible Tech, for their help on our two large-scale human studies. Finally, we thank Dhruba Ghosh, Hannaneh Hajishirzi, and Ludwig Schmidt, the authors of GenEval, for inspiring and supporting our work.  




\clearpage
\newpage
\bibliographystyle{assets/plainnat}
\bibliography{paper}

\begin{thebibliography}{42}
\providecommand{\natexlab}[1]{#1}
\providecommand{\url}[1]{\texttt{#1}}
\expandafter\ifx\csname urlstyle\endcsname\relax
  \providecommand{\doi}[1]{doi: #1}\else
  \providecommand{\doi}{doi: \begingroup \urlstyle{rm}\Url}\fi

\bibitem[Bai et~al.(2025)Bai, Chen, Liu, Wang, Ge, Song, Dang, Wang, Wang, Tang, et~al.]{bai2025qwen2}
Shuai Bai, Keqin Chen, Xuejing Liu, Jialin Wang, Wenbin Ge, Sibo Song, Kai Dang, Peng Wang, Shijie Wang, Jun Tang, et~al.
\newblock Qwen2. 5-vl technical report.
\newblock \emph{arXiv preprint arXiv:2502.13923}, 2025.

\bibitem[Betker et~al.(2023)Betker, Goh, Jing, TimBrooks, Wang, Li, LongOuyang, JuntangZhuang, JoyceLee, YufeiGuo, WesamManassra, PrafullaDhariwal, CaseyChu, YunxinJiao, and Ramesh]{BetkerImprovingIG}
James Betker, Gabriel Goh, Li~Jing, † TimBrooks, Jianfeng Wang, Linjie Li, † LongOuyang, † JuntangZhuang, † JoyceLee, † YufeiGuo, † WesamManassra, † PrafullaDhariwal, † CaseyChu, † YunxinJiao, and Aditya Ramesh.
\newblock Improving image generation with better captions.
\newblock 2023.
\newblock \url{https://api.semanticscholar.org/CorpusID:264403242}.

\bibitem[Chang et~al.(2025)Chang, Fang, Xing, Wu, Cheng, Wang, Zeng, Yu, and Chen]{chang2025oneig}
Jingjing Chang, Yixiao Fang, Peng Xing, Shuhan Wu, Wei Cheng, Rui Wang, Xianfang Zeng, Gang Yu, and Hai-Bao Chen.
\newblock Oneig-bench: Omni-dimensional nuanced evaluation for image generation.
\newblock \emph{arXiv preprint arXiv:2506.07977}, 2025.

\bibitem[Chen et~al.(2025{\natexlab{a}})Chen, Xu, Pan, Hu, Qin, Goldstein, Huang, Zhou, Xie, Savarese, et~al.]{chen2025blip3}
Jiuhai Chen, Zhiyang Xu, Xichen Pan, Yushi Hu, Can Qin, Tom Goldstein, Lifu Huang, Tianyi Zhou, Saining Xie, Silvio Savarese, et~al.
\newblock Blip3-o: A family of fully open unified multimodal models-architecture, training and dataset.
\newblock \emph{arXiv preprint arXiv:2505.09568}, 2025{\natexlab{a}}.

\bibitem[Chen et~al.(2019)Chen, Wang, Pang, Cao, Xiong, Li, Sun, Feng, Liu, Xu, et~al.]{chen2019mmdetection}
Kai Chen, Jiaqi Wang, Jiangmiao Pang, Yuhang Cao, Yu~Xiong, Xiaoxiao Li, Shuyang Sun, Wansen Feng, Ziwei Liu, Jiarui Xu, et~al.
\newblock Mmdetection: Open mmlab detection toolbox and benchmark.
\newblock \emph{arXiv preprint arXiv:1906.07155}, 2019.

\bibitem[Chen et~al.(2025{\natexlab{b}})Chen, Lin, Xu, Shen, Yao, Rimchala, Zhang, and Huang]{chen2025r2i}
Kaijie Chen, Zihao Lin, Zhiyang Xu, Ying Shen, Yuguang Yao, Joy Rimchala, Jiaxin Zhang, and Lifu Huang.
\newblock R2i-bench: Benchmarking reasoning-driven text-to-image generation.
\newblock \emph{arXiv preprint arXiv:2505.23493}, 2025{\natexlab{b}}.

\bibitem[Cheng et~al.(2022{\natexlab{a}})Cheng, Misra, Schwing, Kirillov, and Girdhar]{cheng2022masked}
Bowen Cheng, Ishan Misra, Alexander~G Schwing, Alexander Kirillov, and Rohit Girdhar.
\newblock Masked-attention mask transformer for universal image segmentation.
\newblock In \emph{Proceedings of the IEEE/CVF conference on computer vision and pattern recognition}, pages 1290--1299, 2022{\natexlab{a}}.

\bibitem[Cheng et~al.(2022{\natexlab{b}})Cheng, Misra, Schwing, Kirillov, and Girdhar]{maskformer}
Bowen Cheng, Ishan Misra, Alexander~G. Schwing, Alexander Kirillov, and Rohit Girdhar.
\newblock Masked-attention mask transformer for universal image segmentation.
\newblock In \emph{Proceedings of the IEEE/CVF Conference on Computer Vision and Pattern Recognition (CVPR)}, pages 1290--1299, June 2022{\natexlab{b}}.

\bibitem[Cho et~al.(2023)Cho, Hu, Garg, Anderson, Krishna, Baldridge, Bansal, Pont-Tuset, and Wang]{cho2023davidsonian}
Jaemin Cho, Yushi Hu, Roopal Garg, Peter Anderson, Ranjay Krishna, Jason Baldridge, Mohit Bansal, Jordi Pont-Tuset, and Su~Wang.
\newblock Davidsonian scene graph: Improving reliability in fine-grained evaluation for text-to-image generation.
\newblock \emph{arXiv preprint arXiv:2310.18235}, 2023.

\bibitem[Comanici et~al.(2025)Comanici, Bieber, Schaekermann, Pasupat, Sachdeva, Dhillon, Blistein, Ram, Zhang, Rosen, et~al.]{comanici2025gemini}
Gheorghe Comanici, Eric Bieber, Mike Schaekermann, Ice Pasupat, Noveen Sachdeva, Inderjit Dhillon, Marcel Blistein, Ori Ram, Dan Zhang, Evan Rosen, et~al.
\newblock Gemini 2.5: Pushing the frontier with advanced reasoning, multimodality, long context, and next generation agentic capabilities.
\newblock \emph{arXiv preprint arXiv:2507.06261}, 2025.

\bibitem[Deep-Floyd(2023)]{ifxl2023}
Deep-Floyd.
\newblock Deep-floyd/if.
\newblock 2023.
\newblock \url{https://github.com/deep-floyd/IF}.

\bibitem[Deng et~al.(2025)Deng, Zhu, Li, Gou, Li, Wang, Zhong, Yu, Nie, Song, et~al.]{deng2025emerging}
Chaorui Deng, Deyao Zhu, Kunchang Li, Chenhui Gou, Feng Li, Zeyu Wang, Shu Zhong, Weihao Yu, Xiaonan Nie, Ziang Song, et~al.
\newblock Emerging properties in unified multimodal pretraining.
\newblock \emph{arXiv preprint arXiv:2505.14683}, 2025.

\bibitem[Esser et~al.(2024)Esser, Kulal, Blattmann, Entezari, M{\"u}ller, Saini, Levi, Lorenz, Sauer, Boesel, et~al.]{esser2024scaling}
Patrick Esser, Sumith Kulal, Andreas Blattmann, Rahim Entezari, Jonas M{\"u}ller, Harry Saini, Yam Levi, Dominik Lorenz, Axel Sauer, Frederic Boesel, et~al.
\newblock Scaling rectified flow transformers for high-resolution image synthesis.
\newblock In \emph{Forty-first international conference on machine learning}, 2024.

\bibitem[Ge et~al.(2024)Ge, Zhao, Zhu, Ge, Yi, Song, Li, Ding, and Shan]{ge2024seed}
Yuying Ge, Sijie Zhao, Jinguo Zhu, Yixiao Ge, Kun Yi, Lin Song, Chen Li, Xiaohan Ding, and Ying Shan.
\newblock Seed-x: Multimodal models with unified multi-granularity comprehension and generation.
\newblock \emph{arXiv preprint arXiv:2404.14396}, 2024.

\bibitem[Ghosh et~al.(2023)Ghosh, Hajishirzi, and Schmidt]{ghosh2023geneval}
Dhruba Ghosh, Hannaneh Hajishirzi, and Ludwig Schmidt.
\newblock Geneval: An object-focused framework for evaluating text-to-image alignment.
\newblock \emph{Advances in Neural Information Processing Systems}, 36:\penalty0 52132--52152, 2023.

\bibitem[Hu et~al.(2023)Hu, Liu, Kasai, Wang, Ostendorf, Krishna, and Smith]{hu2023tifa}
Yushi Hu, Benlin Liu, Jungo Kasai, Yizhong Wang, Mari Ostendorf, Ranjay Krishna, and Noah~A Smith.
\newblock Tifa: Accurate and interpretable text-to-image faithfulness evaluation with question answering.
\newblock In \emph{Proceedings of the IEEE/CVF International Conference on Computer Vision}, pages 20406--20417, 2023.

\bibitem[Huang et~al.(2023)Huang, Sun, Xie, Li, and Liu]{huang2023t2i}
Kaiyi Huang, Kaiyue Sun, Enze Xie, Zhenguo Li, and Xihui Liu.
\newblock T2i-compbench: A comprehensive benchmark for open-world compositional text-to-image generation.
\newblock \emph{Advances in Neural Information Processing Systems}, 36:\penalty0 78723--78747, 2023.

\bibitem[Hurst et~al.(2024)Hurst, Lerer, Goucher, Perelman, Ramesh, Clark, Ostrow, Welihinda, Hayes, Radford, et~al.]{hurst2024gpt}
Aaron Hurst, Adam Lerer, Adam~P Goucher, Adam Perelman, Aditya Ramesh, Aidan Clark, AJ~Ostrow, Akila Welihinda, Alan Hayes, Alec Radford, et~al.
\newblock Gpt-4o system card.
\newblock \emph{arXiv preprint arXiv:2410.21276}, 2024.

\bibitem[Invisible(2025)]{invisible}
Invisible.
\newblock Invisible tech.
\newblock 2025.
\newblock \url{https://invisibletech.ai/}.

\bibitem[Labs(2024)]{flux2024}
Black~Forest Labs.
\newblock Flux.
\newblock \url{https://github.com/black-forest-labs/flux}, 2024.

\bibitem[Li et~al.(2024)Li, Lin, Pathak, Li, Fei, Wu, Ling, Xia, Zhang, Neubig, et~al.]{li2024genai}
Baiqi Li, Zhiqiu Lin, Deepak Pathak, Jiayao Li, Yixin Fei, Kewen Wu, Tiffany Ling, Xide Xia, Pengchuan Zhang, Graham Neubig, et~al.
\newblock Genai-bench: Evaluating and improving compositional text-to-visual generation.
\newblock \emph{arXiv preprint arXiv:2406.13743}, 2024.

\bibitem[Lin et~al.(2014)Lin, Maire, Belongie, Hays, Perona, Ramanan, Doll{\'a}r, and Zitnick]{lin2014microsoft}
Tsung-Yi Lin, Michael Maire, Serge Belongie, James Hays, Pietro Perona, Deva Ramanan, Piotr Doll{\'a}r, and C~Lawrence Zitnick.
\newblock Microsoft coco: Common objects in context.
\newblock In \emph{European conference on computer vision}, pages 740--755. Springer, 2014.

\bibitem[Lin et~al.(2024)Lin, Pathak, Li, Li, Xia, Neubig, Zhang, and Ramanan]{lin2024evaluating}
Zhiqiu Lin, Deepak Pathak, Baiqi Li, Jiayao Li, Xide Xia, Graham Neubig, Pengchuan Zhang, and Deva Ramanan.
\newblock Evaluating text-to-visual generation with image-to-text generation.
\newblock In \emph{European Conference on Computer Vision}, pages 366--384. Springer, 2024.

\bibitem[Niu et~al.(2025)Niu, Ning, Zheng, Jin, Lin, Jin, Liao, Feng, Ning, Zhu, et~al.]{niu2025wise}
Yuwei Niu, Munan Ning, Mengren Zheng, Weiyang Jin, Bin Lin, Peng Jin, Jiaqi Liao, Chaoran Feng, Kunpeng Ning, Bin Zhu, et~al.
\newblock Wise: A world knowledge-informed semantic evaluation for text-to-image generation.
\newblock \emph{arXiv preprint arXiv:2503.07265}, 2025.

\bibitem[Pan et~al.(2025)Pan, Shukla, Singh, Zhao, Mishra, Wang, Xu, Chen, Li, Juefei-Xu, et~al.]{pan2025transfer}
Xichen Pan, Satya~Narayan Shukla, Aashu Singh, Zhuokai Zhao, Shlok~Kumar Mishra, Jialiang Wang, Zhiyang Xu, Jiuhai Chen, Kunpeng Li, Felix Juefei-Xu, et~al.
\newblock Transfer between modalities with metaqueries.
\newblock \emph{arXiv preprint arXiv:2504.06256}, 2025.

\bibitem[Peng et~al.(2024)Peng, Cui, Tang, Qi, Dong, Bai, Han, Ge, Zhang, and Xia]{peng2024dreambench++}
Yuang Peng, Yuxin Cui, Haomiao Tang, Zekun Qi, Runpei Dong, Jing Bai, Chunrui Han, Zheng Ge, Xiangyu Zhang, and Shu-Tao Xia.
\newblock Dreambench++: A human-aligned benchmark for personalized image generation.
\newblock \emph{arXiv preprint arXiv:2406.16855}, 2024.

\bibitem[Podell et~al.(2023)Podell, English, Lacey, Blattmann, Dockhorn, Muller, Penna, and Rombach]{Podell2023SDXLIL}
Dustin Podell, Zion English, Kyle Lacey, A.~Blattmann, Tim Dockhorn, Jonas Muller, Joe Penna, and Robin Rombach.
\newblock Sdxl: Improving latent diffusion models for high-resolution image synthesis.
\newblock \emph{ArXiv}, abs/2307.01952, 2023.
\newblock \url{https://api.semanticscholar.org/CorpusID:259341735}.

\bibitem[Radford et~al.(2021)Radford, Kim, Hallacy, Ramesh, Goh, Agarwal, Sastry, Askell, Mishkin, Clark, et~al.]{radford2021learning}
Alec Radford, Jong~Wook Kim, Chris Hallacy, Aditya Ramesh, Gabriel Goh, Sandhini Agarwal, Girish Sastry, Amanda Askell, Pamela Mishkin, Jack Clark, et~al.
\newblock Learning transferable visual models from natural language supervision.
\newblock In \emph{International conference on machine learning}, pages 8748--8763. PmLR, 2021.

\bibitem[Rombach et~al.(2022)Rombach, Blattmann, Lorenz, Esser, and Ommer]{rombach2022high}
Robin Rombach, Andreas Blattmann, Dominik Lorenz, Patrick Esser, and Bj{\"o}rn Ommer.
\newblock High-resolution image synthesis with latent diffusion models.
\newblock In \emph{Proceedings of the IEEE/CVF conference on computer vision and pattern recognition}, pages 10684--10695, 2022.

\bibitem[Schuhmann et~al.(2022)Schuhmann, Beaumont, Vencu, Gordon, Wightman, Cherti, Coombes, Katta, Mullis, Wortsman, et~al.]{schuhmann2022laion}
Christoph Schuhmann, Romain Beaumont, Richard Vencu, Cade Gordon, Ross Wightman, Mehdi Cherti, Theo Coombes, Aarush Katta, Clayton Mullis, Mitchell Wortsman, et~al.
\newblock Laion-5b: An open large-scale dataset for training next generation image-text models.
\newblock \emph{Advances in neural information processing systems}, 35:\penalty0 25278--25294, 2022.

\bibitem[Sun et~al.(2025)Sun, Fang, Duan, Liu, and Liu]{sun2025t2i}
Kaiyue Sun, Rongyao Fang, Chengqi Duan, Xian Liu, and Xihui Liu.
\newblock T2i-reasonbench: Benchmarking reasoning-informed text-to-image generation.
\newblock \emph{arXiv preprint arXiv:2508.17472}, 2025.

\bibitem[Wang et~al.(2024{\natexlab{a}})Wang, Bai, Tan, Wang, Fan, Bai, Chen, Liu, Wang, Ge, et~al.]{wang2024qwen2}
Peng Wang, Shuai Bai, Sinan Tan, Shijie Wang, Zhihao Fan, Jinze Bai, Keqin Chen, Xuejing Liu, Jialin Wang, Wenbin Ge, et~al.
\newblock Qwen2-vl: Enhancing vision-language model's perception of the world at any resolution.
\newblock \emph{arXiv preprint arXiv:2409.12191}, 2024{\natexlab{a}}.

\bibitem[Wang et~al.(2024{\natexlab{b}})Wang, Zhang, Luo, Sun, Cui, Wang, Zhang, Wang, Li, Yu, et~al.]{wang2024emu3}
Xinlong Wang, Xiaosong Zhang, Zhengxiong Luo, Quan Sun, Yufeng Cui, Jinsheng Wang, Fan Zhang, Yueze Wang, Zhen Li, Qiying Yu, et~al.
\newblock Emu3: Next-token prediction is all you need.
\newblock \emph{arXiv preprint arXiv:2409.18869}, 2024{\natexlab{b}}.

\bibitem[Wei et~al.(2025)Wei, Zhang, Wang, Wei, Guo, and Zhang]{wei2025tiif}
Xinyu Wei, Jinrui Zhang, Zeqing Wang, Hongyang Wei, Zhen Guo, and Lei Zhang.
\newblock Tiif-bench: How does your t2i model follow your instructions?
\newblock \emph{arXiv preprint arXiv:2506.02161}, 2025.

\bibitem[Wiles et~al.(2025)Wiles, Zhang, Albuquerque, Kajić, Wang, Bugliarello, Onoe, Papalampidi, Ktena, Knutsen, Rashtchian, Nawalgaria, Pont-Tuset, and Nematzadeh]{wiles2025revisitingtexttoimageevaluationgecko}
Olivia Wiles, Chuhan Zhang, Isabela Albuquerque, Ivana Kajić, Su~Wang, Emanuele Bugliarello, Yasumasa Onoe, Pinelopi Papalampidi, Ira Ktena, Chris Knutsen, Cyrus Rashtchian, Anant Nawalgaria, Jordi Pont-Tuset, and Aida Nematzadeh.
\newblock Revisiting text-to-image evaluation with gecko: On metrics, prompts, and human ratings, 2025.
\newblock \url{https://arxiv.org/abs/2404.16820}.

\bibitem[Wu et~al.(2025{\natexlab{a}})Wu, Li, Zhou, Lin, Gao, Yan, Yin, Bai, Xu, Chen, et~al.]{wu2025qwen}
Chenfei Wu, Jiahao Li, Jingren Zhou, Junyang Lin, Kaiyuan Gao, Kun Yan, Sheng-ming Yin, Shuai Bai, Xiao Xu, Yilei Chen, et~al.
\newblock Qwen-image technical report.
\newblock \emph{arXiv preprint arXiv:2508.02324}, 2025{\natexlab{a}}.

\bibitem[Wu et~al.(2025{\natexlab{b}})Wu, Chen, Wu, Ma, Liu, Pan, Liu, Xie, Yu, Ruan, et~al.]{wu2025janus}
Chengyue Wu, Xiaokang Chen, Zhiyu Wu, Yiyang Ma, Xingchao Liu, Zizheng Pan, Wen Liu, Zhenda Xie, Xingkai Yu, Chong Ruan, et~al.
\newblock Janus: Decoupling visual encoding for unified multimodal understanding and generation.
\newblock In \emph{Proceedings of the Computer Vision and Pattern Recognition Conference}, pages 12966--12977, 2025{\natexlab{b}}.

\bibitem[Xiao et~al.(2025)Xiao, Wang, Zhou, Yuan, Xing, Yan, Li, Wang, Huang, and Liu]{xiao2025omnigen}
Shitao Xiao, Yueze Wang, Junjie Zhou, Huaying Yuan, Xingrun Xing, Ruiran Yan, Chaofan Li, Shuting Wang, Tiejun Huang, and Zheng Liu.
\newblock Omnigen: Unified image generation.
\newblock In \emph{Proceedings of the Computer Vision and Pattern Recognition Conference}, pages 13294--13304, 2025.

\bibitem[Xie et~al.(2024)Xie, Mao, Bai, Zhang, Wang, Lin, Gu, Chen, Yang, and Shou]{xie2024show}
Jinheng Xie, Weijia Mao, Zechen Bai, David~Junhao Zhang, Weihao Wang, Kevin~Qinghong Lin, Yuchao Gu, Zhijie Chen, Zhenheng Yang, and Mike~Zheng Shou.
\newblock Show-o: One single transformer to unify multimodal understanding and generation.
\newblock \emph{arXiv preprint arXiv:2408.12528}, 2024.

\bibitem[Yang et~al.(2025)Yang, Li, Yang, Zhang, Hui, Zheng, Yu, Gao, Huang, Lv, et~al.]{yang2025qwen3}
An~Yang, Anfeng Li, Baosong Yang, Beichen Zhang, Binyuan Hui, Bo~Zheng, Bowen Yu, Chang Gao, Chengen Huang, Chenxu Lv, et~al.
\newblock Qwen3 technical report.
\newblock \emph{arXiv preprint arXiv:2505.09388}, 2025.

\bibitem[Ye et~al.(2025)Ye, Jiang, Wang, Zhu, Hu, Huang, He, Yan, Yu, Li, et~al.]{ye2025echo}
Junyan Ye, Dongzhi Jiang, Zihao Wang, Leqi Zhu, Zhenghao Hu, Zilong Huang, Jun He, Zhiyuan Yan, Jinghua Yu, Hongsheng Li, et~al.
\newblock Echo-4o: Harnessing the power of gpt-4o synthetic images for improved image generation.
\newblock \emph{arXiv preprint arXiv:2508.09987}, 2025.

\bibitem[Zhou et~al.(2024)Zhou, Yu, Babu, Tirumala, Yasunaga, Shamis, Kahn, Ma, Zettlemoyer, and Levy]{zhou2024transfusion}
Chunting Zhou, Lili Yu, Arun Babu, Kushal Tirumala, Michihiro Yasunaga, Leonid Shamis, Jacob Kahn, Xuezhe Ma, Luke Zettlemoyer, and Omer Levy.
\newblock Transfusion: Predict the next token and diffuse images with one multi-modal model.
\newblock \emph{arXiv preprint arXiv:2408.11039}, 2024.

\end{thebibliography}

\clearpage
\newpage
\beginappendix

\section{Example of a Failure Case of the GenEval Detector}
\label{sec:app_failure_case}
\begin{figure}[h]
    \centering
    \includegraphics[width=\linewidth]{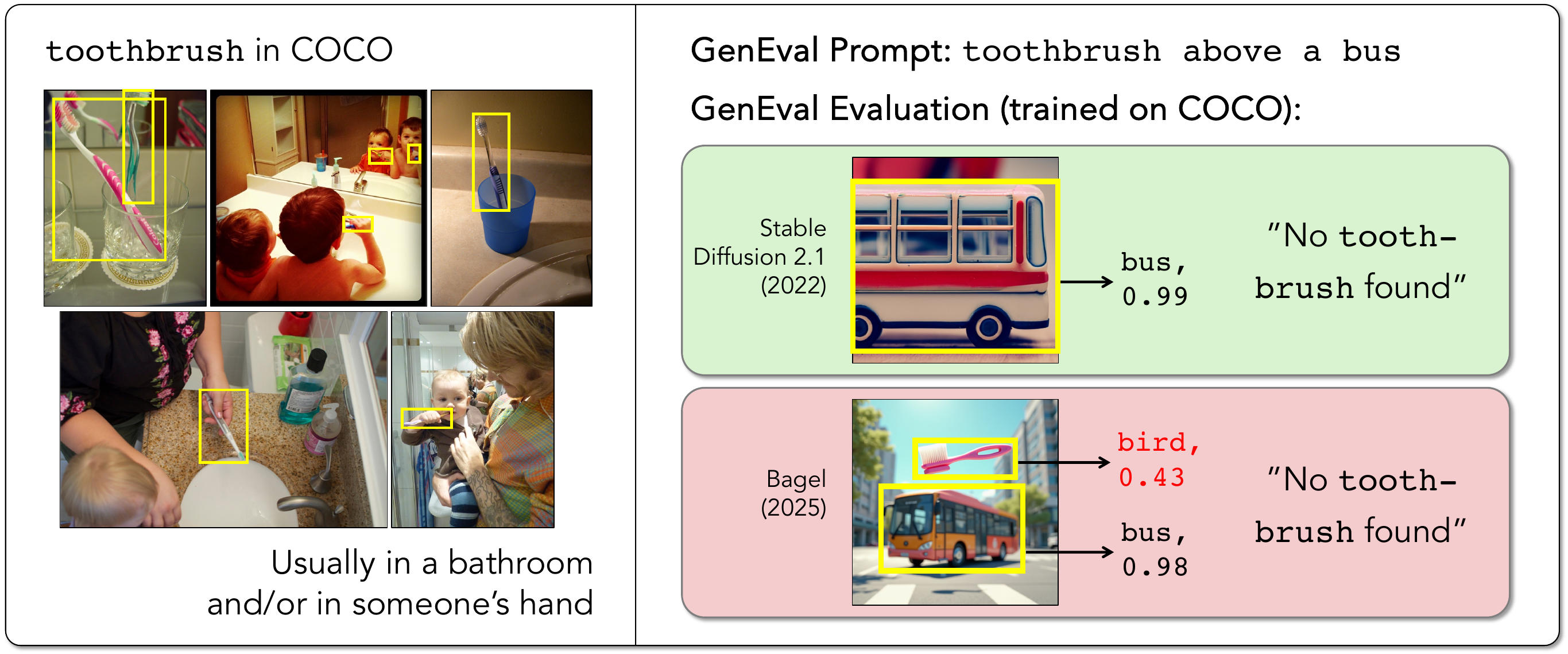}
    \caption{The object detector in GenEval fails to recognize the (correctly placed) toothbrush in the image generated by Bagel, mislabeling it as a bird: likely because in the COCO dataset the object detector was trained on, toothbrushes tend to appear in bathrooms and hands, while birds tend to appear in the sky. This results in an incorrect evaluation of the image correctly generated by Bagel.}
    \label{fig:teaser_appendix}
\end{figure}

Figure \ref{fig:teaser_appendix} shows a second failure case of the GenEval evaluation rooted in its object detector, after the example from Figure \ref{fig:teaser}. As the T2I output distribution shifted farther from COCO with the advent of higher scales of training (including on synthetic data), the object detector in GenEval trained on COCO breaks down. 

\section{\benchmark~Curation Details}
\label{sec:app_ge2_creation}
\subsection{Full Vocabulary of \benchmark}
\label{sec:app_ge2_vocab}
The prompts in \benchmark~consist of words randomly sampled from the vocabulary below. The objects sourced from COCO are in \textbf{bold}.

\paragraph{Animate objects:} \texttt{\textbf{bear}, \textbf{bird}, \textbf{cat}, \textbf{cow}, \textbf{dog}, \textbf{elephant}, flamingo, \textbf{giraffe}, \textbf{horse}, kangaroo, koala, lion, monkey, penguin, pig, rabbit, raccoon, \textbf{sheep}, turtle, \textbf{zebra}}. 

\paragraph{Inanimate objects:} \texttt{\textbf{backpack}, bagel, \textbf{bicycle}, candle, \textbf{car}, \textbf{chair}, \textbf{clock}, cookie, croissant, \textbf{donut}, flower, guitar, \textbf{motorcycle}, mushroom, \textbf{suitcase}, toy, \textbf{truck}, trumpet, \textbf{umbrella}, violin}.

\paragraph{Attributes:} \texttt{red, yellow, green, blue, purple, pink, brown, black, white, spotted, striped, checkered, sparkling, wooden, glass, plastic, metal, stone}.

\paragraph{Prepositions:} \texttt{to the left of, to the right of, on top of, under, in front of, behind}.

\paragraph{Transitive Verb Relations:} \texttt{chasing, playing with, jumping over}.

\paragraph{Counts:} \texttt{two, three, four, five, six, seven}.

\subsection{GPT-4o Rewriting Prompt}
\label{sec:app_ge2_rewriting}
In order to rewrite \benchmark~prompts in a longer, more detailed format, in-line with training and evaluation methods of recent T2I models, we use the prompt provided in Figure \ref{fig:app_rewriting_prompt}. The in-context examples are GenEval prompts alongside their rewritten equivalents as provided in the BAGEL repository\footnote{\url{https://github.com/bytedance-seed/BAGEL}}. We select 110 words as the suggested maximum as 550 of 553 rewritten GenEval prompts are within that length.

\begin{figure}[h]
\centering
\fbox{%
  \begin{minipage}{0.95\linewidth}
I am evaluating a text-to-image generation model. Given a short prompt, rewrite the prompt to a more detailed version.
    
Here are three examples:

Prompt: a photo of a dog right of a teddy bear

Detailed Prompt: A soft, beige teddy bear sits upright on a wooden floor. The teddy bear has round black eyes and a small stitched nose. Its fur is slightly matted, giving it a well-loved appearance. To the right of the teddy bear, a golden retriever dog stands on all fours. The dog's fur is smooth and shiny, catching the light. Its ears are perked up, and its tail is slightly raised. The dog's eyes are focused forward, exuding a calm and attentive expression. The scene is set in a realistic, photographic style, with natural lighting highlighting the textures of both the teddy bear and the dog. \\

Prompt: a photo of two backpacks

Detailed Prompt: There are two backpacks placed side by side on a wooden floor. Both backpacks are medium-sized and upright. The first backpack is black with a sleek, smooth texture. The second backpack is navy blue with a slightly rugged appearance. Each backpack has multiple zippered compartments visible on the front. The straps of both backpacks rest neatly against their sides. The lighting is soft, highlighting the realistic textures and colors of the materials. The scene is simple, focusing only on the two backpacks in a photographic style. \\

Prompt: a photo of a green bus and a purple microwave

Detailed Prompt: A vibrant green bus is parked on the left side of the scene. The bus has a glossy, metallic finish that reflects the surrounding light realistically. Its large windows are clean and transparent, revealing empty seats inside. On the right side, a compact purple microwave sits on a flat surface. The microwave has a matte finish, with a small digital display and a silver handle on its front. Both objects are positioned without overlap, ensuring their distinct colors and details are clearly visible. The scene is rendered in a photographic style, emphasizing lifelike textures and realistic lighting. \\

Here is the prompt for you to rewrite:

Prompt: \{\}

Detailed Prompt:

Reply with the detailed prompt alone, with no extra text. Make sure the detailed prompt is under 110 words.
  \end{minipage}
}
\caption{Prompt given to GPT-4o to rewrite \benchmark.}
\label{fig:app_rewriting_prompt}
\end{figure}

\section{User Study Details}
\label{sec:app_user_study}
In this section, we provide the instructions provided to annotators for the two user studies in our work, alongside the user interface shown to the annotators. 

\begin{figure}
    \centering
    \includegraphics[width=0.5\linewidth]{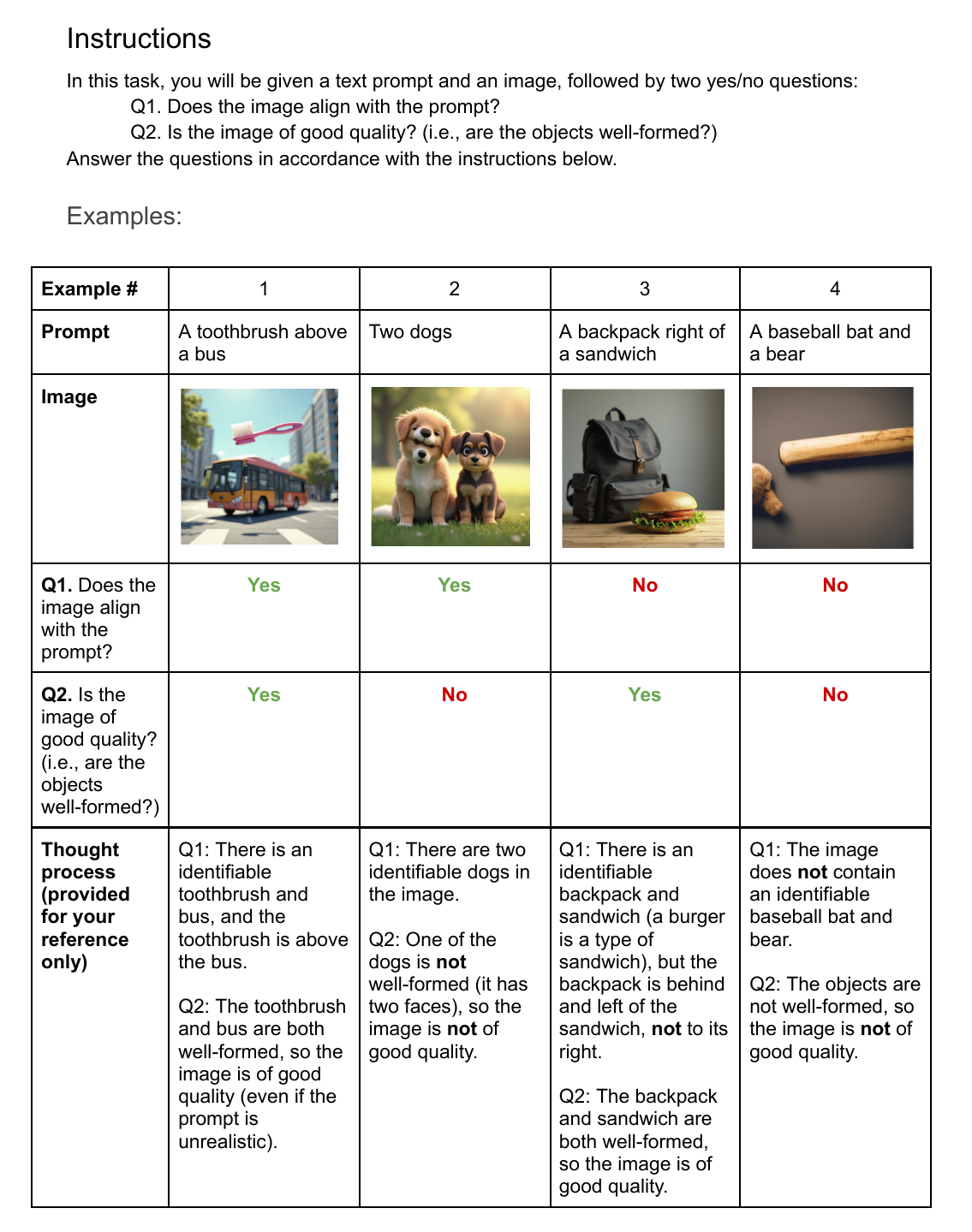}
    \caption{Annotator Instructions for GenEval User Study: Part 1}
    \label{fig:app_ge1_instructions_1}
\end{figure}

\begin{figure}
    \centering
    \includegraphics[width=0.5\linewidth]{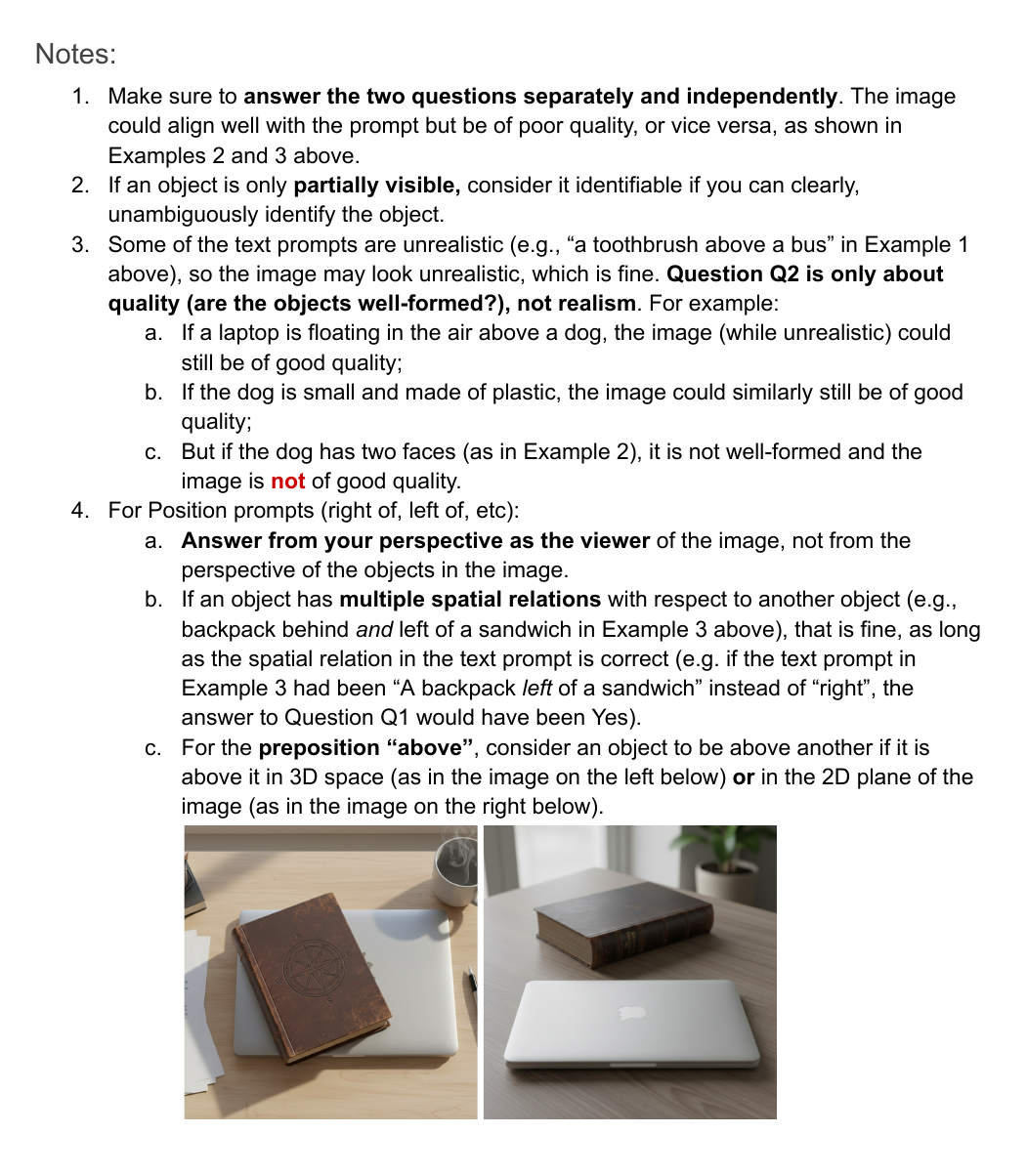}
    \caption{Annotator Instructions for GenEval User Study: Part 2}
    \label{fig:app_ge1_instructions_2}
\end{figure}

\subsection{GenEval User Study}
\label{sec:app_ge1_user_study}
\begin{figure}
    \centering
    \includegraphics[width=0.5\linewidth]{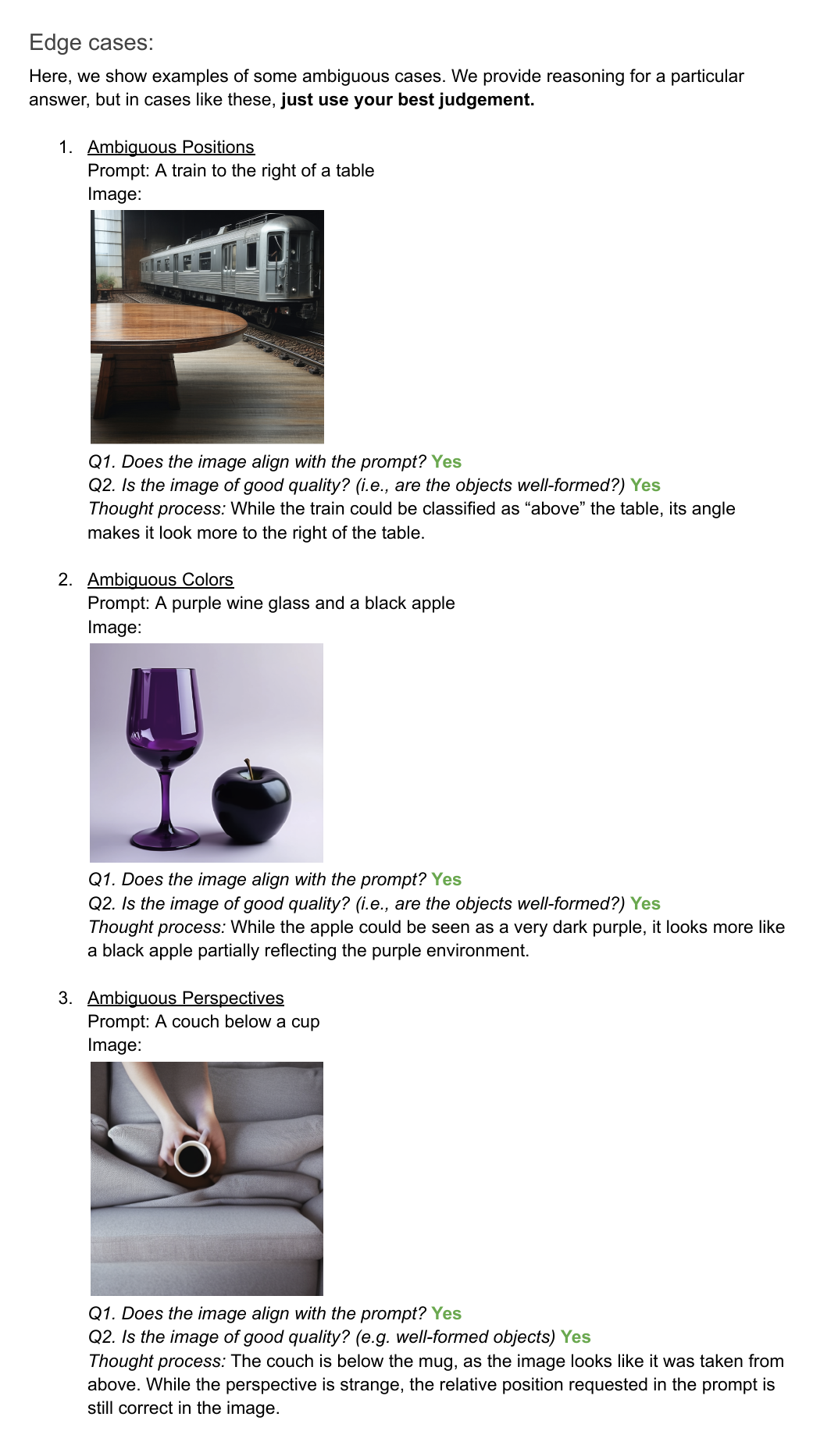}
    \caption{Annotator Instructions for GenEval User Study: Part 3}
    \label{fig:app_ge1_instructions_3}
\end{figure}

For this study, annotators were shown an image and two yes/no questions: (1) ``Does the image align with the prompt?''; and (2) ``Is the image of good quality? (i.e., are the objects well-formed?)''. The full instructions shown to the annotators are provided in Figures \ref{fig:app_ge1_instructions_1}, \ref{fig:app_ge1_instructions_2} and \ref{fig:app_ge1_instructions_3}.

\subsection{\benchmark~User Study}
\label{sec:app_ge2_user_study}
\begin{figure}[h]
    \centering
    \includegraphics[width=0.5\linewidth]{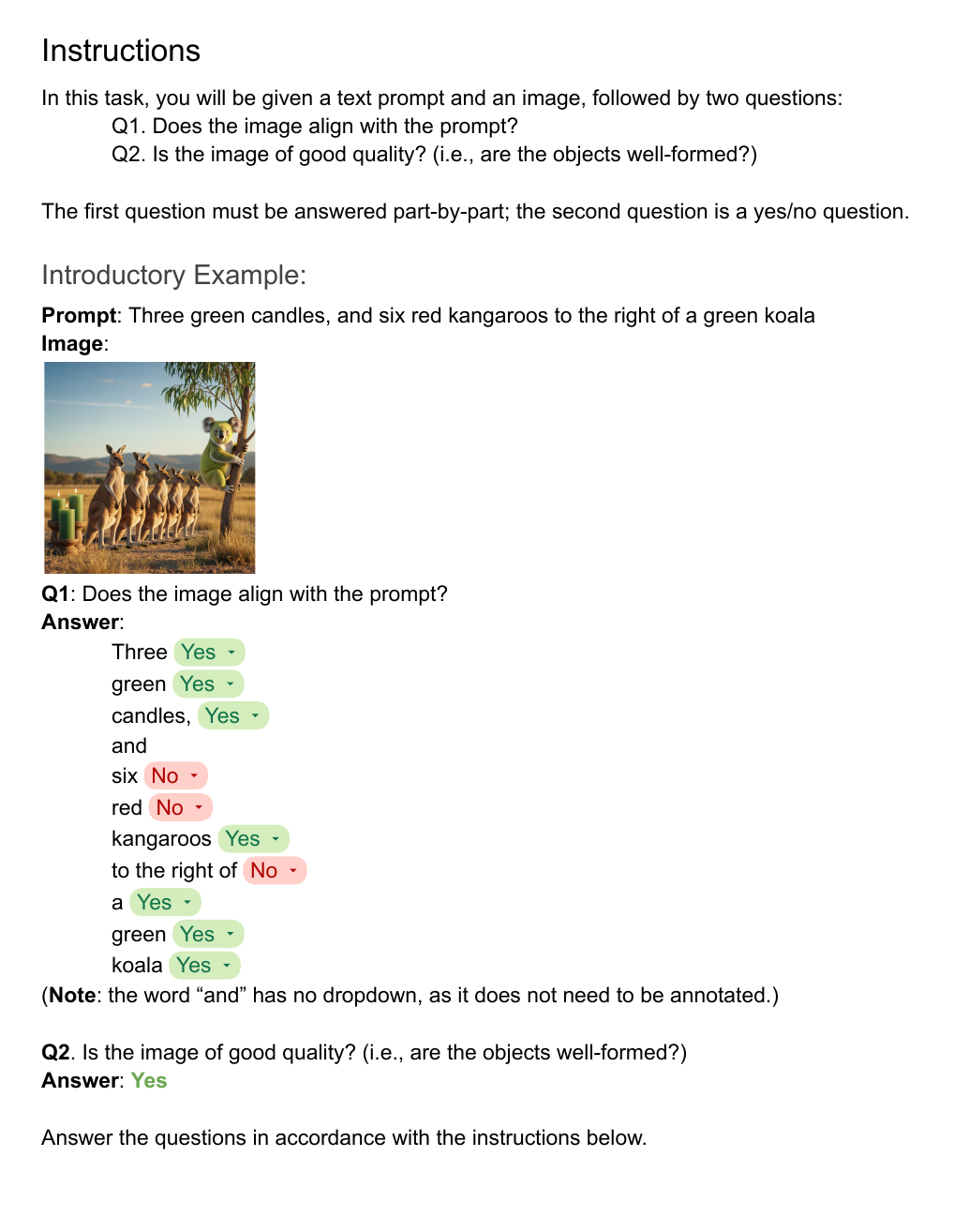}
    \caption{Annotator Instructions for \benchmark~Study}
    \label{fig:app_ge2_instructions_1}
\end{figure}

\begin{figure}[h]
    \centering
    \includegraphics[width=0.5\linewidth]{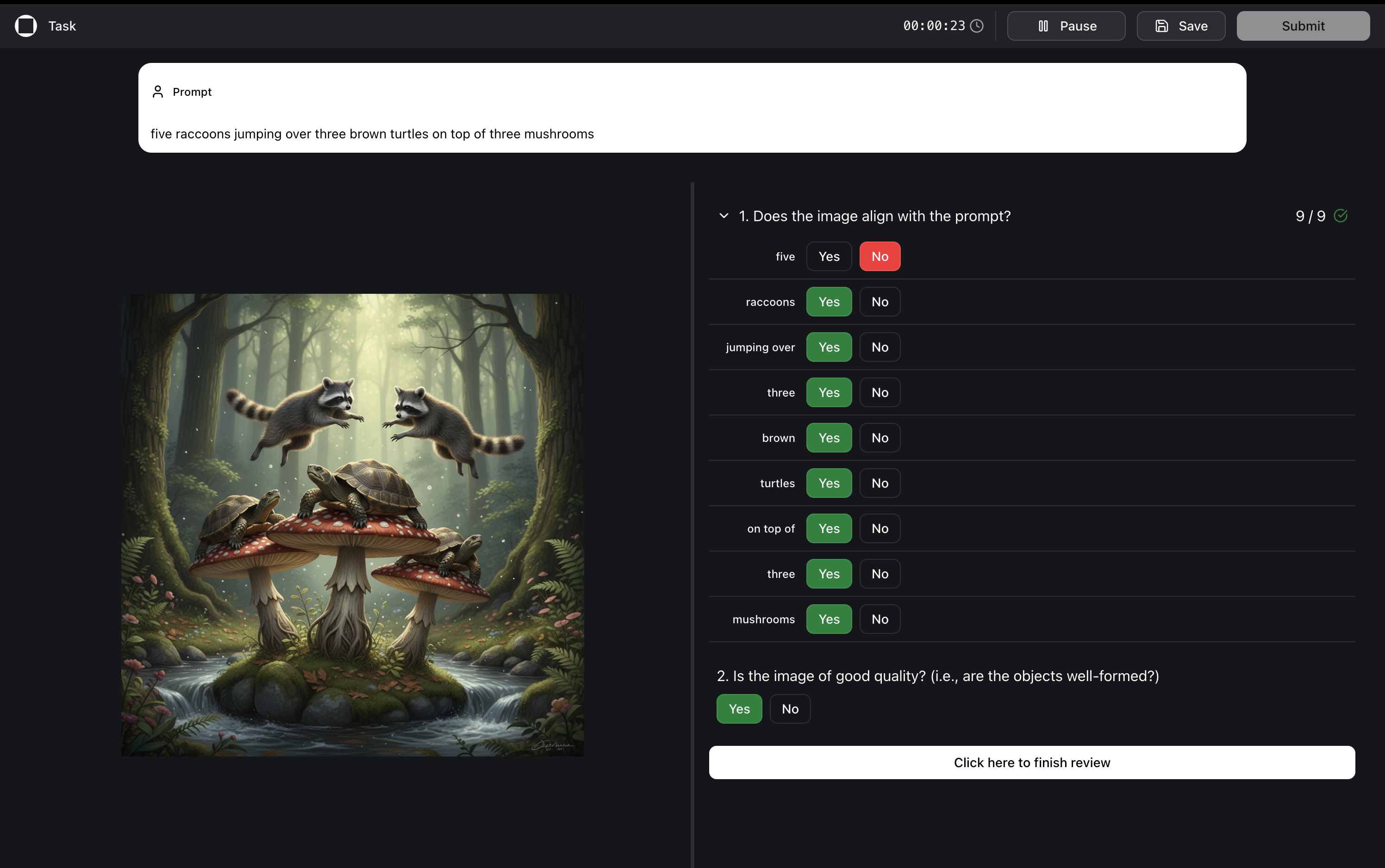}
    \caption{User Interface for \benchmark~User Study}
    \label{fig:app_ge2_user_interface}
\end{figure}

For this user study, we follow the same structure as the user study on GenEval, but instead of asking a yes/no question for ``Does the image align with the prompt'', we provide a checklist for each atom in the prompt for the user to select based on alignment of the image. Thus, we obtain atom-level annotations of correctness for each model. The instructions shown to the annotators are provided in Figure \ref{fig:app_ge2_instructions_1}, with the remainder of the instructions similar to the user study for GenEval (but with per-atom annotations in the examples).
The user interface shown to the annotators is provided in Figure \ref{fig:app_ge2_user_interface}. The user interface for the GenEval study was similar, but with a single yes/no question for alignment rather than per-atom annotations.

\section{Per-Model Analyses of Human-Judged Scores on \benchmark}
\label{sec:app_per_model}
In Section \ref{sec:geneval_2}, we discuss and analyze human-judged scores on \benchmark~averaged over 4 state-of-the-art T2I models: Flux.1-dev, Bagel, Qwen-Image and Gemini 2.5 Flash. Here, we present results for each of the 8 T2I models. 

In Table \ref{tab:app_per_model_skill}, we show how each model performs on each skill evaluated in \benchmark. Various inferences can be drawn: for example, Gemini struggles the most with counting, achieving only 69.2\% accuracy. Meanwhile, Qwen-Image struggles the most on relations between objects (both spatial and transitive verb), achieving 61.1\% accuracy on each. These analyses allow targeted diagnosis and improvement of T2I model flaws.

In Table \ref{tab:app_per_atomicity}, we present results for each model at each level of prompt atomicity, i.e., compositionality, or complexity. Unsurprisingly, all models uniformly perform much better on less complex prompts than on more complex ones. However, this drop is quite sharp for many models: e.g., Flux.1-dev achieves 25\% accuracy at atomicity=6, and 0\% accuracy at all higher atomicities, potentially shedding light on the complexity of its training data. \benchmark~highlights the significant room for improvement of all T2I models at higher compositionality.

\begin{table*}[h]
\centering
\begin{tabular}{l c c c c c}
\toprule
& \multicolumn{5}{c}{\textbf{Skill}}\\
\textbf{Model} & \textbf{Object} & \textbf{Attribute} & \textbf{Count} & \textbf{Position} & \textbf{Verb} \\
\midrule
 \multicolumn{1}{l}{\textit{Stable Diffusion Model Series}} &  &\\
SD2.1  & 55.1 & 30.4 & 22.3 & 11.7 & 17.8 \\
SDXL & 74.1 &	42.4	&28.7	&16.0	&28.9 \\
SD3 & 87.0	& 65.5	& 49.9	& 41.0	& 46.7 \\
SD3.5-large & 91.6	& 70.3 & 52.2& 	39.5& 	55.6 \\
\midrule
 \multicolumn{1}{l}{\textit{State-of-the-Art T2I Models}} & &\\
Flux & 88.4	& 68.3	& 55.6	& 37.0	& 44.4 \\
Bagel+CoT & 92.9	& 75.9	& 55.6	& 50.6	& 57.8 \\
Qwen & 99.1	& 85.6	& 70.3	& 60.2	& 71.1\\
Gemini & 99.0	& 91.4	& 70.1	& 70.2	& 86.7 \\
\bottomrule
\end{tabular}
\caption{Per-model performance across skills on \benchmark. T2I models particularly struggle with counting, spatial relations, and transitive verb relations.}
\label{tab:app_per_model_skill}
\end{table*}

\begin{table*}[h]
\centering
\begin{tabular}{l c c c c c c c c}
\toprule
& \multicolumn{8}{c}{\textbf{Atomicity}}\\
\textbf{Model} & \textbf{3} & \textbf{4} & \textbf{5} & \textbf{6} & \textbf{7} & \textbf{8} & \textbf{9} & \textbf{10} \\
\midrule
 \multicolumn{1}{l}{\textit{Stable Diffusion Model Series}} &  &\\
SD2.1  & 12.0	& 8.0	& 0.0	& 0.0	& 0.0	& 0.0	& 0.0	& 0.0 \\
SDXL   & 24.0	& 8.0	& 2.0	& 0.0	& 0.0	& 0.0	& 0.0	& 0.0 \\
SD3    & 52.0	& 30.0	& 16.0	& 8.0	& 0.0	& 4.0	& 2.0	& 0.0 \\
SD3.5-large  & 46.0	& 34.0	& 12.0	& 14.0	& 4.0	& 2.0	& 2.0	& 0.0 \\
\midrule
 \multicolumn{1}{l}{\textit{State-of-the-Art T2I Models}} & &\\
Flux   & 48.0	& 28.0	& 16.0	& 26.0	& 4.0	& 0.0	& 0.0	& 0.0 \\
Bagel+CoT  & 46.0	& 34.0	& 18.0	& 14.0	& 8.0	& 6.0	& 0.0 & 2.0 \\
Qwen   & 58.0	& 46.0	& 40.0	& 34.0	& 10.0	& 20.0	& 4.0	& 2.0\\
Gemini & 58.0	& 58.0	& 42.0	& 36.0	& 18.0	& 20.0	& 12.0	& 4.0 \\
\bottomrule
\end{tabular}
\caption{Per-model performance across varying levels of prompt compositionality (atomicity) on \benchmark. T2I model performance drops sharply as prompt compositionality increases.}
\label{tab:app_per_atomicity}
\end{table*}

\clearpage
\section{Illustrative Example of Soft-TIFA}
\label{sec:app_soft_tifa_example}
\begin{figure}[h]
    \centering
    \includegraphics[width=\linewidth]{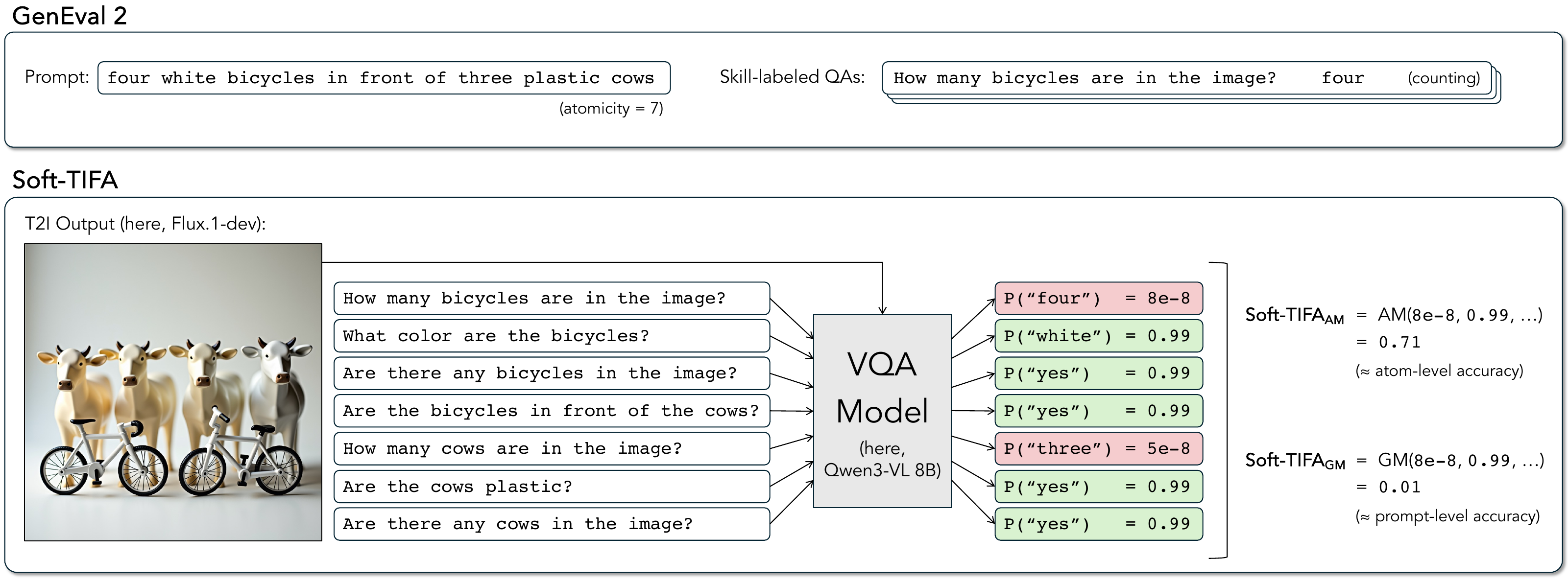}
    \caption{An example from \benchmark, alongside its evaluation with Soft-TIFA. \benchmark~examples are annotated with atomicity and skills, allowing for detailed analyses. \am~provides an estimate of atom-level T2I performance, while \gm~provides an estimate of prompt-level T2I performance.}
    \label{fig:app_soft_tifa}
\end{figure}

Figure \ref{fig:app_soft_tifa} shows an example from the \benchmark~benchmark, alongside the Soft-TIFA evaluation method. Soft-TIFA has the advantages of both soft scores (as in VQAScore) as well as per-atom questions (as in TIFA), while attaining full coverage over the prompt due to the templated nature of \benchmark. 

\section{Evaluation Methods with Different Underlying VQA Models}
\label{sec:app_different_vqa}
We present in Table \ref{tab:app_all_qwens} the AUROC of VQAScore, TIFA, \am~and \gm~for different types of prompt sets (original, rewritten, and both) of \benchmark, with different underlying VQA models: Qwen3-VL, Qwen2.5-VL, Qwen2-VL and GPT-4o. \gm~has higher human alignment than all other evaluation methods on \benchmark~under all versions of Qwen. While VQAScore outperforms \gm~with GPT-4o as the VQA model, the absolute AUROC is significantly lower than methods using Qwen, with Qwen3-VL providing the highest human alignment. It is further worth noting that as GPT-4o is closed-source, it is unclear exactly what is being returned as the ``logprobs'' of various output tokens.

\begin{table*}[h]
\centering
\resizebox{0.66\linewidth}{!}{%
\begin{tabular}{l l c c c c}
\toprule
\textbf{Model} & \textbf{Prompt} & \textbf{VQAScore} & \textbf{TIFA} &
\textbf{\am} & \textbf{\gm} \\
\midrule
Qwen3-VL   & Original  & \underline{93.5} & 91.2 & 92.8 & \textbf{94.2} \\
        & Rewritten & 90.5 & 91.6 & \underline{92.9} & \textbf{94.4} \\
        & Both      & 92.4 & 91.6 & \underline{93.0} & \textbf{94.5} \\
\midrule
Qwen2.5-VL & Original  & 89.8 & 90.5 & \underline{92.0} & \textbf{92.9} \\
        & Rewritten & 85.2 & 88.8 & \underline{90.8} & \textbf{92.6} \\
        & Both      & 88.1 & 90.0 & \underline{91.7} & \textbf{93.0} \\
\midrule
Qwen2-VL   & Original  & 89.1 & 90.7 & \underline{92.4} & \textbf{93.1} \\
        & Rewritten & 83.7 & 88.2 & \underline{91.4} & \textbf{92.5} \\
        & Both      & 87.0 & 89.8 & \underline{92.2} & \textbf{93.0} \\
\midrule
GPT-4o  & Original  & \textbf{92.1} & 88.2 & 89.8 & \underline{91.7} \\
        & Rewritten & \textbf{89.6} & 83.9 & 86.0 & \underline{88.8} \\
        & Both      & \textbf{91.2} & 86.6 & 88.4 & \underline{90.6} \\
\bottomrule
\end{tabular}
}
\caption{AUROC of VQAScore, TIFA, \am~and \gm~across different prompt types, with different underlying VQA models. \gm~outperforms all other methods with all Qwen models. Although GPT-4o VQAScore outperforms GPT-4o \gm, it is significantly lower than \gm~with Qwen models. In each row, the highest AUROC is in \textbf{bold} and the second-highest is \underline{underlined}.
}
\label{tab:app_all_qwens}
\end{table*}

\section{Per-Skill and Per-Compositionality Analyses of Soft-TIFA}
\label{sec:app_soft_tifa_per}
In this section, we present the per-skill and per-compositionality performance of state-of-the-art (SOTA) T2I models (Flux.1-dev, Bagel, Qwen-Image and Gemini 2.5 Flash), as estimated by Soft-TIFA. As skill is an atom-level concept and compositionality is a prompt-level concept, we use \am~to estimate T2I performance on the former, and \gm~for the latter. 

Note that as Soft-TIFA scores are soft rather than binary (as the human-judged scores are), these results are overall slightly regularized compared to those in Section \ref{sec:ge2_analyses}; i.e., high scores are slightly lower (e.g., where Soft-TIFA assigned a score of 0.99 instead of 1), and low scores are slightly higher (e.g., where Soft-TIFA assigned a score of 0.01 instead of 0). We focus on the trends the results follow.

\begin{figure*}[h]
    \centering
    \begin{subfigure}[h]{0.48\linewidth}
        \centering
        \includegraphics[width=0.9\linewidth]{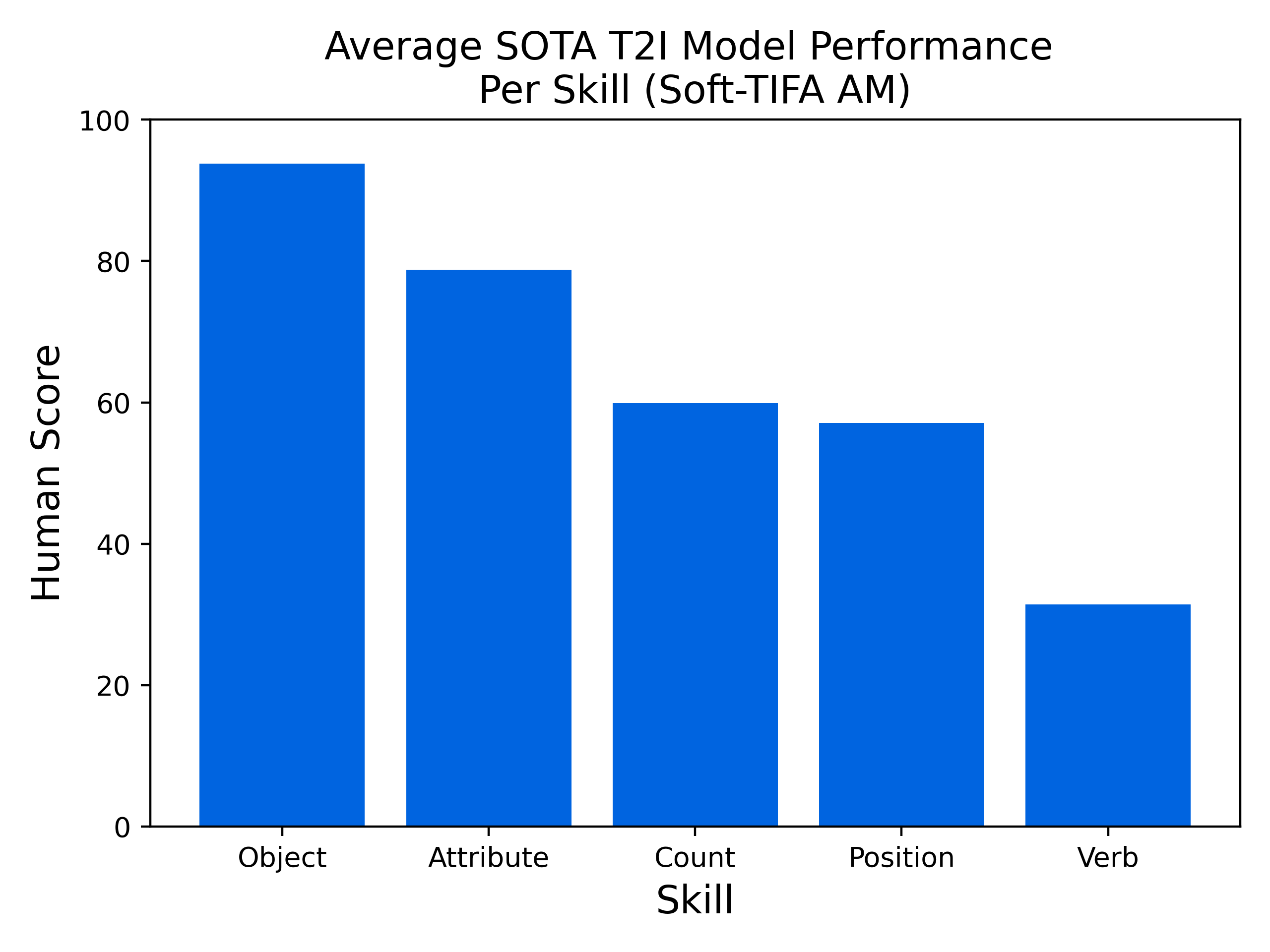}
        \caption{Average \am~scores of SOTA T2I models per skill in \benchmark.}
        \label{fig:geneval2_skill_soft_tifa}
    \end{subfigure}
    \hfill
    \begin{subfigure}[h]{0.48\linewidth}
        \centering
        \includegraphics[width=0.9\linewidth]{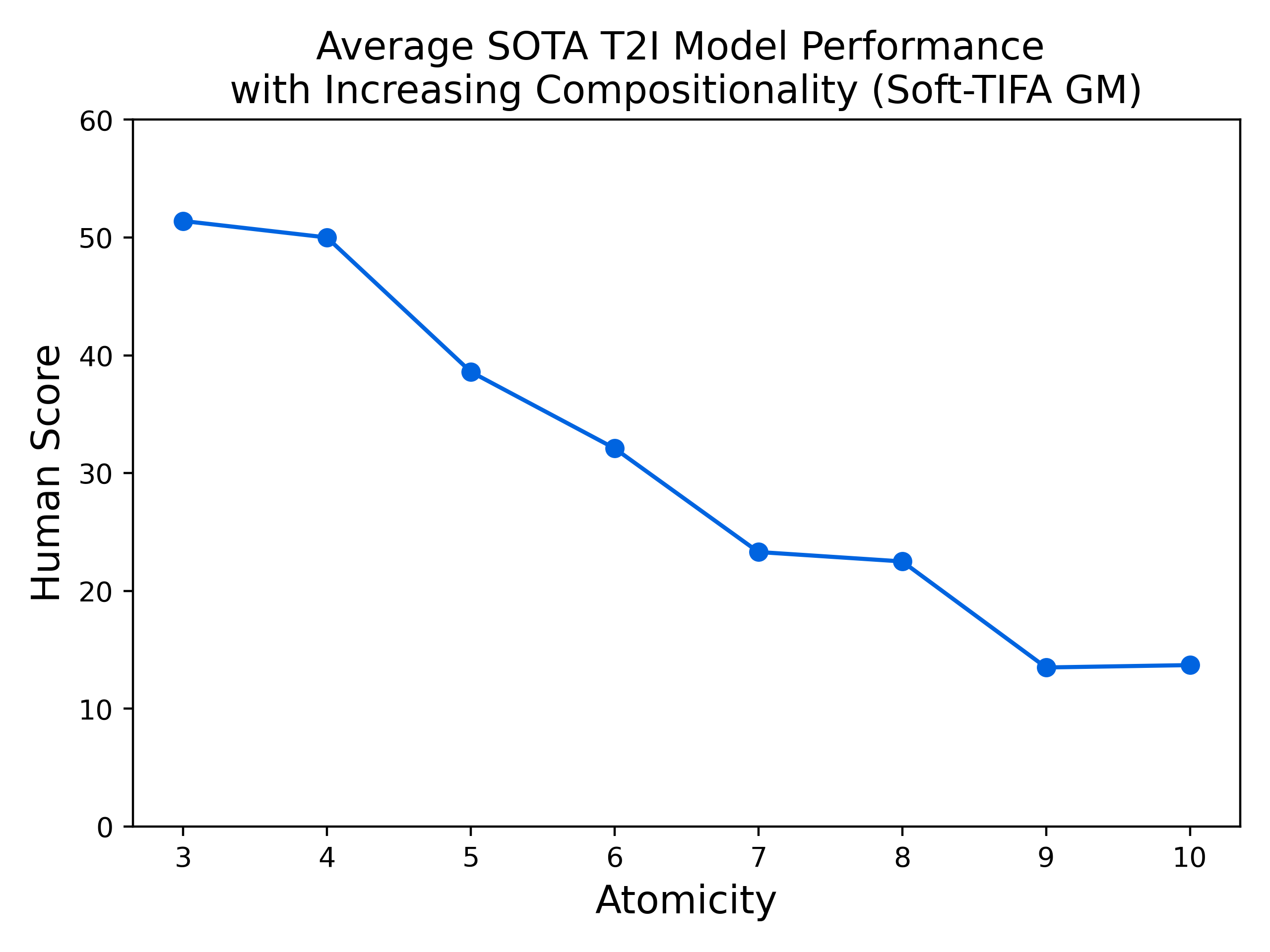}
        \caption{Average \gm~scores of SOTA T2I models for each level of compositionality (``atomicity'') in \benchmark.}
        \label{fig:geneval2_compositionality_soft_tifa}
    \end{subfigure}
    \caption{
    \benchmark~enables various analyses of T2I models: (a) atom-level analyses such as performance per-skill with \am; and (b) prompt-level analyses such as performance per-compositionality with \gm. The trends of Soft-TIFA reflect the human-judged scores from our user study (Figure \ref{fig:two_side_by_side}), but for the ``verb'' skill, in which VQA models are stricter than human annotators.}
    \label{fig:app_soft_tifa_per_skill_comp}
\end{figure*}

Our results are presented in Figure \ref{fig:app_soft_tifa_per_skill_comp}. The trends largely mirror the human-judged scores collected in our large-scale user study and shown in Figure \ref{fig:two_side_by_side}, but for the ``verb'' skill, in which the VQA model is significantly stricter than the human annotators, awarding a score of $31.4$ compared to human judgment at $65.0$. While the fact that the majority of the Soft-TIFA trends mirror human judgment shows promise for Soft-TIFA as a T2I evaluation method, this finding also serves as a reminder that VQA-based methods are only as strong as their underlying VQA models are, and calls for further research in visual understanding of more nuanced relations between objects.
\vspace{1em}

\end{document}